\documentclass{elsarticle}
\usepackage{amsmath}
\usepackage{amssymb}
\usepackage{booktabs}
\usepackage{comment}
\usepackage{xcolor}
  \usepackage{xcolor}
  \usepackage{tikz}
\providecolor{causecol}{RGB}{214,234,248}
\providecolor{detcol}{RGB}{220,240,220}
\providecolor{linecol}{RGB}{90,90,90}
\providecolor{edgeone}{RGB}{192,57,43}
\providecolor{edgetwo}{RGB}{230,126,34}
\providecolor{edgethree}{RGB}{142,68,173}
\providecolor{edgefour}{RGB}{22,160,133}
\providecolor{edgefive}{RGB}{121,85,72}
 
\providecommand{\flink}[5]{\draw[#1] ($(#2.east)+(0,#3)$) to[out=0,in=180] ($(#4.west)+(0,#5)$);}
  \usetikzlibrary{arrows.meta,calc,fit,backgrounds}
\usepackage{tikz}
\usepackage{tikz}
\usetikzlibrary{arrows.meta,positioning,shapes.geometric}
\usepackage{colortbl}
\newcolumntype{L}[1]{>{\raggedright\arraybackslash}p{#1}}
\usetikzlibrary{arrows.meta,positioning,shapes.geometric}
\usepackage{booktabs}
\usepackage{array}
\usepackage{graphicx}
\usepackage{tikz}
\usetikzlibrary{arrows.meta,positioning,shapes.geometric}
\usepackage{booktabs}
\usepackage{array}
\usepackage{colortbl}
\usepackage{graphicx}
\usepackage{enumitem}

\usepackage{array}
\usepackage{longtable}
\usepackage{graphicx}
\usepackage{enumitem}
\usepackage[hidelinks]{hyperref}
\usepackage{xcolor}
\usepackage{float}
\usepackage{url}
\usepackage{graphicx}

\journal{Journal Name}

\begin{document}

\begin{frontmatter}

\title{Plausible but Ungrounded: Etiology, Detection, and Verification Limits of LLM Hallucination}

\author[inst1]{Hillsea Rana}
\ead{rhillsea@gmail.com}
\author[inst1]{Amit Bhola}
\ead{amit.bhola@sharda.ac.in}

\author[inst1]{Piyush Singh}
\ead{2023362845.piyush@ug.sharda.ac.in}

\author[inst1]{Yash Kumar}
\ead{2023340572.yash@ug.sharda.ac.in}

\affiliation[inst1]{
    organization={Department of Computer Science and Engineering, Sharda University},
    city={Greater Noida},
    country={India}
}

\begin{abstract}

Large language models (LLMs) write fluently on almost any topic, and that fluency is what makes their mistakes dangerous: hallucinated content reads as convincingly as correct content, and users often cannot tell the difference. This review explains why LLMs hallucinate and how such errors can be caught. The paper first brings together different taxonomies into a multidimensional framework that incorporates perspectives based on sources (both intrinsic and extrinsic), factuality, and faithfulness, along with some finer aspects such as FRANK, HILT, orientation, modality, and severity. It subsequently describes the origins of hallucination like noisy training data, static parametric knowledge, objectives that prefer likelihood and usefulness, variance in prompts, errors in retrieval, and low visual grounding in multimodal models, and explains its aggravation through such specific areas as medicine, programming, and spatial planning.  Additionally, it compares the characteristics of different detection methods, starting from human judgment and reference-based NLI approaches to claim verification, token and semantic entropy, self-consistency, LLM-as-a-Judge, and specialized multimodal detection techniques. Finally, the paper identifies five major research gaps: hallucination origin attribution, calibrated uncertainty, generalization across domains and tasks, multilingual and multimodal robustness, and the reliability of verification methods. Researchers and practitioners building or deploying trustworthy LLM systems will find a map of what is known, what each detector can and cannot see, and what remains unsolved.

\end{abstract}


\end{frontmatter}

\section{Introduction}
\label{sec:intro}

Large language models (LLMs) write fluently on almost any topic, and that fluency is exactly what makes their mistakes dangerous. \textit{Hallucination} refers to generated content that is fluent and syntactically correct but factually incorrect, unsupported by the available context, or inconsistent with the instructions and evidence given to the model\cite{AnhHoang2025Survey,Augenstein2023Factuality}. The outcome may seem credible but is actually fake, covering all possible topics. It should also be noted that since LLMs do not operate in a closed environment and beyond a certain task, hallucination is a concept relating to numerous failures that may include: factual inaccuracies, groundless claims, contextual or reasoning errors In the area of medicine for example, even a minor mistake can have fatal consequences \cite{Augenstein2023Factuality}.While many advances have been made in learning how to spot a hallucination, still there definite challenges to face \cite{Asgari2025ClinicalSafety}. It has to be noted that new challenges relate not only to hallucination itself but also to its detection. Each missing element of the study weakens one detection factor, while the area of detection lacks information about the previous one .Dashed blue arrows show which causes each detector must cope with; solid colored arrows show which detector limitations give rise to each gap, and the color of an arrow matches the color of its gap.

\begin{figure}[H]
\centering
\resizebox{\textwidth}{!}{%
\begin{tikzpicture}[
  font=\footnotesize,
  box/.style={rectangle, rounded corners=3pt, minimum height=1.0cm,
              text width=4.5cm, align=center, inner sep=2pt, line width=0.7pt},
  cause/.style={box, draw=blue!55!black, fill=causecol},
  amp/.style={box, draw=blue!55!black, fill=causecol!40, dashed},
  det/.style={box, draw=green!40!black, fill=detcol},
  hyb/.style={box, draw=green!40!black, fill=green!25, line width=1.1pt},
  gap/.style={box, text width=5.0cm, minimum height=1.3cm, line width=1.2pt},
  hdr/.style={rectangle, rounded corners=3pt, minimum height=0.8cm,
              text=white, font=\bfseries\small, align=center, inner sep=3pt},
  grp/.style={rectangle, rounded corners=5pt, draw=gray!70, dashed,
              inner sep=6pt, line width=0.6pt},
  grplab/.style={font=\scriptsize\bfseries\itshape, text=gray!60!black,
                 inner sep=1.5pt},
  cdedge/.style={-{Stealth[length=1.4mm]}, blue!50!black!70, dashed,
                 line width=0.5pt},
  ef1/.style={-{Stealth[length=1.7mm]}, edgeone,   line width=0.9pt},
  ef2/.style={-{Stealth[length=1.7mm]}, edgetwo,   line width=0.9pt},
  ef3/.style={-{Stealth[length=1.7mm]}, edgethree, line width=0.9pt},
  ef4/.style={-{Stealth[length=1.7mm]}, edgefour,  line width=0.9pt},
  ef5/.style={-{Stealth[length=1.7mm]}, edgefive,  line width=0.9pt}
]
 
\node[hdr, fill=blue!55!black,  text width=4.5cm] (HA) at (0,1.9)    {Why it happens\\[-1pt]{\scriptsize Section 2: causes}};
\node[hdr, fill=green!40!black, text width=4.5cm] (HB) at (7.2,1.9)  {How it is caught\\[-1pt]{\scriptsize Section 3: detection}};
\node[hdr, fill=orange!70!black,text width=5.0cm] (HC) at (14.8,1.9) {What remains open\\[-1pt]{\scriptsize research gaps G1--G5}};
\draw[-{Stealth[length=3mm]}, line width=1.2pt, gray] (HA.east) -- (HB.west);
\draw[-{Stealth[length=3mm]}, line width=1.2pt, gray] (HB.east) -- (HC.west);
 
\node[cause] (A1) at (0,0)    {\textbf{Data and corpus}\\[-1pt]{\scriptsize noisy, biased, outdated}};
\node[cause] (A2) at (0,-1.2) {\textbf{Knowledge limits}\\[-1pt]{\scriptsize static, incomplete knowledge}};
\node[cause] (A3) at (0,-2.4) {\textbf{Optimization, alignment}\\[-1pt]{\scriptsize likelihood, not truth}};
\node[cause] (A4) at (0,-4.4) {\textbf{Prompt and decoding}\\[-1pt]{\scriptsize ambiguity, sampling}};
\node[cause] (A5) at (0,-5.6) {\textbf{Retrieval and context}\\[-1pt]{\scriptsize missed, weak, ignored evidence}};
\node[cause] (A6) at (0,-6.8) {\textbf{Multimodal grounding}\\[-1pt]{\scriptsize language priors beat the image}};
\node[amp]   (A7) at (0,-8.8) {\textbf{Domain, task amplifiers}\\[-1pt]{\scriptsize healthcare, code, planning}};
 
\node[det] (B1) at (7.2,0)    {\textbf{Human evaluation}\\[-1pt]{\scriptsize validation standard, costly}};
\node[det] (B2) at (7.2,-1.2) {\textbf{Reference-based, NLI}\\[-1pt]{\scriptsize entailed or contradicted}};
\node[det] (B3) at (7.2,-2.4) {\textbf{Claim-level checking}\\[-1pt]{\scriptsize extract, retrieve, verify}};
\node[det] (B4) at (7.2,-4.4) {\textbf{Token, semantic entropy}\\[-1pt]{\scriptsize uncertainty signals}};
\node[det] (B5) at (7.2,-5.6) {\textbf{Self-consistency}\\[-1pt]{\scriptsize stability across samples}};
\node[det] (B6) at (7.2,-7.4) {\textbf{LLM-as-a-Judge}\\[-1pt]{\scriptsize scalable, needs validation}};
\node[det] (B7) at (7.2,-8.6) {\textbf{Visual, multilingual}\\[-1pt]{\scriptsize Woodpecker, ALTI+}};
\node[hyb] (B8) at (7.2,-10.6) {\textbf{Hybrid pipeline}\\[-1pt]{\scriptsize flag, verify, hand off}};
 
\node[gap, draw=edgeone,   fill=edgeone!8]   (G1) at (14.8,-1.0)  {\textbf{G1\; Origin attribution}\\[-1pt]{\scriptsize which stage, mechanism, severity}};
\node[gap, draw=edgetwo,   fill=edgetwo!8]   (G2) at (14.8,-3.4)  {\textbf{G2\; Calibrated uncertainty}\\[-1pt]{\scriptsize confident and systematic errors}};
\node[gap, draw=edgethree, fill=edgethree!8] (G3) at (14.8,-5.8)  {\textbf{G3\; Domain generalization}\\[-1pt]{\scriptsize tasks, long context, code, clinical}};
\node[gap, draw=edgefour,  fill=edgefour!8]  (G4) at (14.8,-8.2)  {\textbf{G4\; Language, modality}\\[-1pt]{\scriptsize low-resource, visual, joint}};
\node[gap, draw=edgefive,  fill=edgefive!8]  (G5) at (14.8,-10.6) {\textbf{G5\; Reliable verifiers}\\[-1pt]{\scriptsize judge bias, stale evidence}};
 
\begin{scope}[on background layer]
  \node[grp, fill=causecol!25, fit=(A1)(A2)(A3), label={[grplab]above:Training-time causes}] {};
  \node[grp, fill=causecol!25, fit=(A4)(A5)(A6), label={[grplab]above:Inference-time causes}] {};
  \node[grp, fill=detcol!35,   fit=(B1)(B2)(B3), label={[grplab]above:Reference-based}] {};
  \node[grp, fill=detcol!35,   fit=(B4)(B5),     label={[grplab]above:Reference-free}] {};
  \node[grp, fill=detcol!35,   fit=(B6)(B7),     label={[grplab]above:Model-based and specialized}] {};
\end{scope}
 
\flink{cdedge}{A1}{0.15}{B2}{0.15}
\flink{cdedge}{A1}{-0.15}{B5}{0.15}
\flink{cdedge}{A2}{0.15}{B2}{-0.15}
\flink{cdedge}{A2}{-0.15}{B3}{0.15}
\flink{cdedge}{A3}{0}{B4}{0.15}
\flink{cdedge}{A4}{0.15}{B4}{-0.15}
\flink{cdedge}{A4}{-0.15}{B5}{-0.15}
\flink{cdedge}{A5}{0}{B3}{-0.15}
\flink{cdedge}{A6}{0}{B7}{0}
\flink{cdedge}{A7}{0}{B8}{0}
 
\flink{ef1}{B3}{0.2}{G1}{0.25}
\flink{ef1}{B8}{0.3}{G1}{-0.25}
\flink{ef2}{B4}{0.2}{G2}{0.2}
\flink{ef2}{B5}{0}{G2}{-0.2}
\flink{ef3}{B1}{0}{G3}{0.3}
\flink{ef3}{B4}{-0.2}{G3}{0}
\flink{ef3}{B8}{0}{G3}{-0.3}
\flink{ef4}{B6}{0.2}{G4}{0.2}
\flink{ef4}{B7}{0}{G4}{-0.2}
\flink{ef5}{B2}{0}{G5}{0.45}
\flink{ef5}{B3}{-0.2}{G5}{0.15}
\flink{ef5}{B6}{-0.2}{G5}{-0.15}
\flink{ef5}{B8}{-0.3}{G5}{-0.45}
 
\node[box, text width=19.3cm, minimum height=0.9cm, draw=gray!70, fill=gray!12]
  at (7.55,-12.7)
  {\textbf{Cross-cutting dimensions:}\; error type (intrinsic, extrinsic, contradiction, fabrication)
   $\cdot$ modality $\cdot$ language $\cdot$ severity $\cdot$ domain risk};
\end{tikzpicture}}
\caption{Roadmap of the review. Left: the causes of hallucination across the
training-time and inference-time pipeline (Section~2). Center: the detection
families (Section~3), grouped by the evidence they use. Right: the five open
research gaps introduced in this section and revisited in the Conclusion. Dashed
blue arrows indicate which causes a detector must cope with; solid colored
arrows indicate which detector limitations give rise to each gap (arrow color =
gap color).}
\label{fig:gap-landscape}
\end{figure}
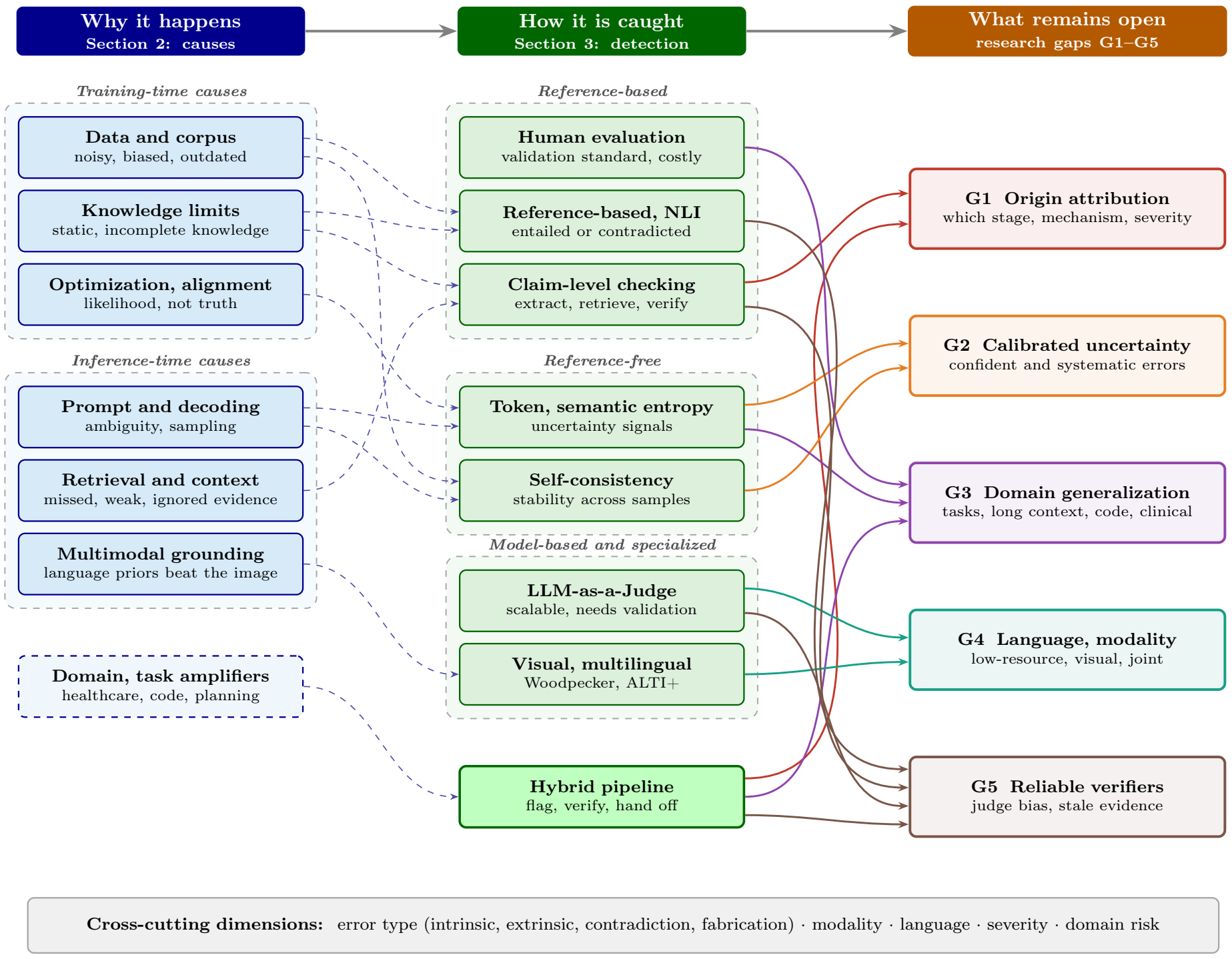

Understanding this family requires answering four questions in turn: how the problem changed over time, how its variants can be told apart, why it arises, and what can be done about it. 
\subsection{From Obvious Nonsense to Convincing Errors}
\label{subsec:history}

The word itself comes from psychology, where it denotes perceiving things that do not exist. Early NLP systems, such as n-gram models (which predict text from counts of short word sequences) and early neural sequence generators, produced mostly nonsensical or repetitive text, so their hallucinations were easy to spot \cite{Ji2023Survey}. Neural translation, summarization, and dialogue systems shifted attention to output judged against a given input, which produced the first systematic classification of the problem \cite{AnhHoang2025Survey}.
Unlike earlier systems that rely primarily on the input, foundation models also draw on \textit{parametric knowledge}, which consists of information absorbed into the model's weights during large-scale pretraining. This allows them to generate responses across open domains. Their errors are therefore fluent and convincing, and much harder for users to detect \cite{Rawte2023FoundationModelSurvey}. Multimodal LLMs, which process images and other inputs alongside text, extended the problem further, to mismatches between generated text and visual evidence \cite{Bai2024MLLMHallucinationSurvey}. Recent work has become more theoretical: some analyses argue that hallucination is inherent to sufficiently expressive, Turing-complete models (models capable of general computation), so it may never be fully eliminated
, while others note that imaginative generation can be valuable in creative uses such as poetry and storytelling \cite{Bai2024MLLMHallucinationSurvey}.

\subsection{Telling Hallucinations Apart}
\label{subsec:taxonomy}

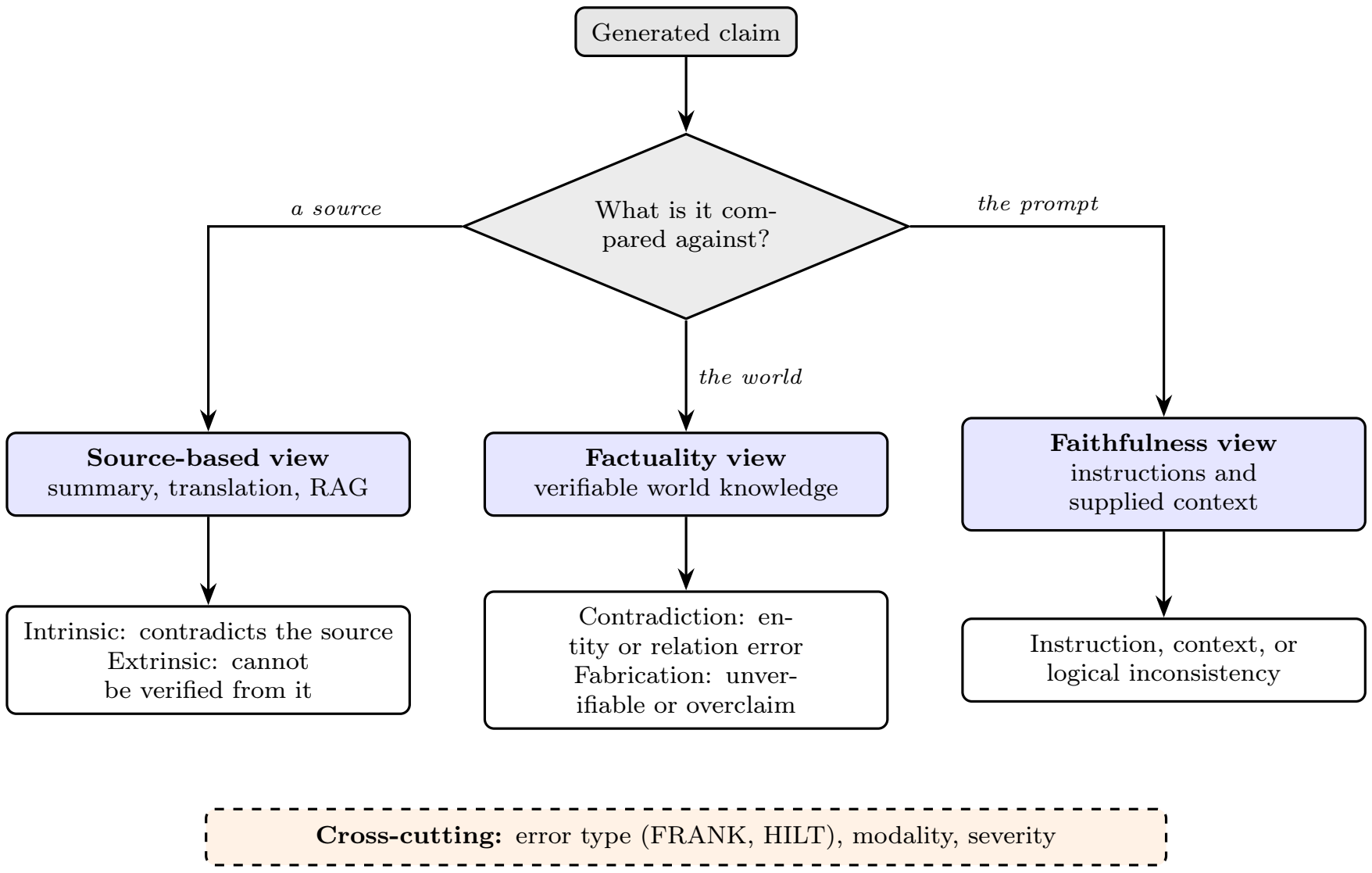
\begin{figure}[H]
\centering
\resizebox{0.95\linewidth}{!}{%
\begin{tikzpicture}[
  n/.style={rectangle, rounded corners=3pt, draw, thick, align=center, font=\footnotesize, inner sep=5pt},
  ref/.style={n, fill=blue!10, text width=4.2cm},
  res/.style={n, fill=white, text width=4.2cm},
  q/.style={diamond, aspect=2.4, draw, thick, fill=gray!15, align=center, font=\footnotesize, inner sep=1pt, text width=3.4cm},
  arr/.style={-{Stealth[length=2.5mm]}, thick},
  lab/.style={font=\scriptsize\itshape}
]
\node[n, fill=gray!20] (o) at (0,0) {Generated claim};
\node[q] (d) at (0,-2.2) {What is it compared against?};
\node[ref] (a) at (-5.4,-5.0) {\textbf{Source-based view}\\ summary, translation, RAG};
\node[ref] (b) at (0,-5.0) {\textbf{Factuality view}\\ verifiable world knowledge};
\node[ref] (c) at (5.4,-5.0) {\textbf{Faithfulness view}\\ instructions and supplied context};
\node[res] (ra) at (-5.4,-7.1) {Intrinsic: contradicts the source\\ Extrinsic: cannot be verified from it};
\node[res] (rb) at (0,-7.1) {Contradiction: entity or relation error\\ Fabrication: unverifiable or overclaim};
\node[res] (rc) at (5.4,-7.1) {Instruction, context, or logical inconsistency};
\node[n, dashed, fill=orange!10, text width=10.5cm] (x) at (0,-9.1) {\textbf{Cross-cutting:} error type (FRANK, HILT), modality, severity};
\draw[arr] (o)--(d);
\draw[arr] (d.west) -| (a.north) node[pos=0.25, above, lab] {a source};
\draw[arr] (d.south) -- (b.north) node[midway, right, lab] {the world};
\draw[arr] (d.east) -| (c.north) node[pos=0.25, above, lab] {the prompt};
\draw[arr] (a)--(ra); \draw[arr] (b)--(rb); \draw[arr] (c)--(rc);
\end{tikzpicture}}
\caption{Decision flow for classifying a hallucination by its reference point. The three views are complementary, and error type, modality, and severity apply across all of them.}
\label{fig:decision}
\end{figure}

Because hallucinations are judged \emph{against something}, the most useful way to classify them is by what the output is compared with. Figure~\ref{fig:decision} turns the literature's main schemes into one decision flow.

The \emph{source-based} view arose in summarization and translation, and it also fits retrieval-augmented generation (RAG), where a model answers from documents fetched at query time. An \textit{intrinsic} hallucination contradicts the source (for instance, stating 2021 when the source says a vaccine was approved in 2019 \cite{Maynez2020FaithfulnessFactuality}), whereas an \textit{extrinsic} one adds content the source cannot confirm, which counts as hallucinated even if it happens to be true \cite{Huang2025Survey}. For general-purpose LLMs, Huang et al.\ separated two further notions \cite{Huang2025Survey}. \textit{Factuality} asks whether the content is true of the real world: it fails through a \textit{contradiction} (a wrong entity such as the wrong person or place, or a wrong relation between correct entities) or a \textit{fabrication} (a claim that cannot be verified at all, or an opinion presented as fact). \textit{Faithfulness} asks whether the response stays consistent with what the model was given: it fails when instructions are ignored, when the supplied context is contradicted, or when the reasoning contradicts itself. The two can diverge: a response may be factually correct yet unfaithful to a ``summarize only this document'' instruction, or faithful to a flawed context while still being wrong. They also demand different verification, either external trusted knowledge or the input and retrieved evidence.

\begin{table}[H]
\centering
\small
\renewcommand{\arraystretch}{1.25}
\caption{Categories of hallucination and their issues}
\label{tab:schemes}
\begin{tabular}{|p{0.20\linewidth}|p{0.22\linewidth}|p{0.46\linewidth}|}
\hline
\rowcolor{gray!15}
\textbf{View} & \textbf{Category} & \textbf{What goes wrong} \\
\hline
Source-based & Intrinsic & Output directly contradicts the source text \\
\hline
Source-based & Extrinsic & Output adds content that cannot be verified from the source \\
\hline
Factuality & Contradiction & Wrong entity, or wrong relation between correct entities \\
\hline
Factuality & Fabrication & Unverifiable claims, or opinion presented as fact \\
\hline
Faithfulness & Instruction & Explicit user instructions are ignored \\
\hline
Faithfulness & Context & Supplied input or documents are ignored or contradicted \\
\hline
Faithfulness & Logical & Self-contradiction, or conclusions that do not follow \\
\hline
\end{tabular}
\end{table}
\subsection{Finer Lenses on the Same Problem}

Binary signs do not reveal what went wrong, thus more precise systems clarify the whole situation. (Table~\ref{tab:fine}). \textbf{FRANK} annotates factual errors in summaries with seven error types at three levels: inside a statement (a wrong predicate, entity, or circumstance such as time, place, or quantity), across statements (a wrong pronoun reference or causal or temporal link), and verifiability (content absent from the source, or grammar so flawed it obscures meaning). Annotators agreed strongly (Cohen's $\kappa = 0.86$) \cite{pagnoni2021understanding}, and its out-of-article category corresponds to extrinsic hallucination \cite{pagnoni2021understanding,laban2022summac}. \textbf{HILT} (HallucInation eLiciTation) instead classifies the \emph{content} of the error into six textual classes, such as \textit{Numeric Nuisance} (wrong dates or amounts), \textit{Generated Golem} (invented people or organizations), \textit{Virtual Voice} (fabricated quotations), and \textit{Time Wrap} (mixed periods) \cite{Rawte2023FoundationModelSurvey}.

Three further lenses cut across both: \textit{orientation}, which asks whether the prompt itself was sound (a \textit{factual mirage} is a wrong output from a correct prompt, a \textit{silver lining} an elaborate narrative built on a misleading one) \cite{Rawte2023FoundationModelSurvey}; \textit{modality}, which in multimodal settings asks whether the wrong object, attribute, or relationship was reported \cite{Bai2024MLLMHallucinationSurvey}; and \textit{severity}, which grades errors from mild to alarming by their deviation from reality and potential harm \cite{Rawte2023FoundationModelSurvey}.

Hence, hallucination is multidimensional rather than binary: a full description specifies the relation to the source, factuality and faithfulness, error type, modality, and severity. The term also hides different mechanisms, since \textit{confabulation} (arbitrary answers that change from run to run) and \textit{systematic} error (consistently repeated because it was learned) have different causes and cures \cite{Farquhar2024Detecting}.

\begin{table}[H]
\centering
\small
\renewcommand{\arraystretch}{1.3}
\caption{The first two lenses classify the error itself; the last three describe dimensions that apply to any error.}
\label{tab:fine}
\begin{tabular}{|>{\raggedright\arraybackslash}p{0.14\linewidth}|>{\raggedright\arraybackslash}p{0.23\linewidth}|>{\raggedright\arraybackslash}p{0.54\linewidth}|}
\hline
\rowcolor{gray!25}
\textbf{Lens} & \textbf{Question it asks} & \textbf{Categories (with plain-language meaning)} \\
\hline
\multicolumn{3}{|l|}{\cellcolor{blue!10}\textbf{A. Lenses that classify the error itself}} \\
\hline
\textbf{FRANK} \cite{pagnoni2021understanding}
& What kind of factual error is in the summary? Seven types in three groups.
& \textit{Within a statement:} wrong predicate (action or relation), wrong entity (person, place, or thing), wrong circumstance (time, place, or quantity). \newline
\textit{Across statements:} wrong coreference (a pronoun points to the wrong entity), wrong discourse link (a false causal or temporal connection). \newline
\textit{Verifiability:} out-of-article content (absent from the source; corresponds to extrinsic hallucination), grammatical error (so flawed that the meaning is lost). \\
\hline
\textbf{HILT} \cite{Rawte2023FoundationModelSurvey}
& What does the false content look like? Six textual classes.
& \textit{Numeric Nuisance} (wrong dates, ages, or amounts); \textit{Acronym Ambiguity} (an acronym misread or wrongly expanded); \textit{Generated Golem} (invented people or organizations); \textit{Virtual Voice} (fabricated quotations); \textit{Geographic Erratum} (wrong place or location); \textit{Time Wrap} (different periods mixed together). \\
\hline
\multicolumn{3}{|l|}{\cellcolor{blue!10}\textbf{B. Cross-cutting lenses that apply to any error}} \\
\hline
\textbf{Orientation} \cite{Rawte2023FoundationModelSurvey}
& Was the prompt itself sound?
& \textit{Factual mirage:} a correct prompt, yet a wrong output. \newline
\textit{Silver lining:} a misleading prompt, answered with an elaborate and convincing narrative. \\
\hline
\textbf{Multimodal} \cite{Bai2024MLLMHallucinationSurvey}
& Does the generated text match what the image shows?
& \textit{Category} error (wrong object reported); \textit{attribute} error (wrong property of a real object, e.g.\ color or size); \textit{relation} error (wrong relationship between objects). \\
\hline
\textbf{Severity} \cite{Rawte2023FoundationModelSurvey}
& How far does the output deviate from reality, and how much harm can it cause?
& \textit{Mild} (minor deviation, low harm); \textit{moderate}; \textit{alarming} (large deviation or serious potential harm). \\
\hline
\end{tabular}
\end{table}

\subsection{Why Hallucinations Arise}
\label{subsec:why}

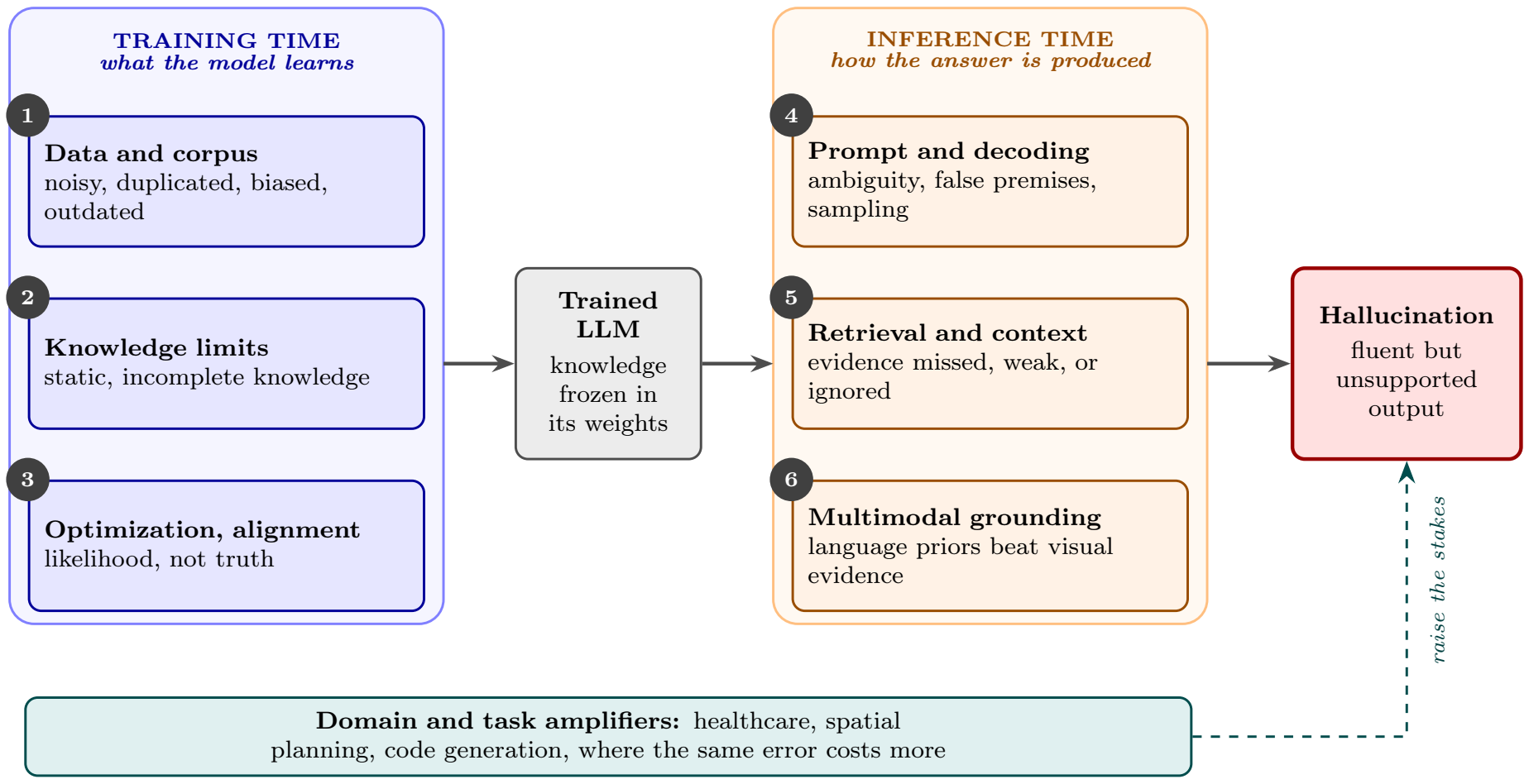
\begin{figure}[H]
\centering
\resizebox{\linewidth}{!}{%
\begin{tikzpicture}[
  cause/.style={rectangle, rounded corners=4pt, draw=#1!60!black, thick, fill=#1!10, align=left, text width=4.2cm, minimum height=1.5cm, inner sep=5pt, font=\footnotesize},
  hub/.style={rectangle, rounded corners=4pt, draw=black!70, thick, fill=gray!15, align=center, text width=1.9cm, minimum height=2.2cm, font=\footnotesize},
  outn/.style={rectangle, rounded corners=4pt, draw=red!60!black, very thick, fill=red!12, align=center, text width=2.4cm, minimum height=2.2cm, font=\footnotesize},
  badge/.style={circle, fill=black!75, text=white, inner sep=1.2pt, minimum size=0.5cm, font=\scriptsize\bfseries},
  band/.style={rounded corners=8pt, thick, draw=#1!50},
  bandlab/.style={font=\scriptsize\bfseries, text=#1!60!black, align=center},
  arr/.style={-{Stealth[length=2.6mm]}, very thick, black!70}
]
\hyphenpenalty=10000 \exhyphenpenalty=10000
\filldraw[band=blue,   fill=blue!4]   (0,-0.5)   rectangle (5.0,6.6);
\filldraw[band=orange, fill=orange!5] (8.8,-0.5) rectangle (13.8,6.6);
\node[bandlab=blue]   at (2.5,6.1)  {TRAINING TIME\\[-1pt]\itshape what the model learns};
\node[bandlab=orange] at (11.3,6.1) {INFERENCE TIME\\[-1pt]\itshape how the answer is produced};
\node[cause=blue] (c1) at (2.5,4.6) {\textbf{Data and corpus}\\ noisy, duplicated, biased, outdated};
\node[cause=blue] (c2) at (2.5,2.5) {\textbf{Knowledge limits}\\ static, incomplete knowledge};
\node[cause=blue] (c3) at (2.5,0.4) {\textbf{Optimization, alignment}\\ likelihood, not truth};
\node[cause=orange] (c4) at (11.3,4.6) {\textbf{Prompt and decoding}\\ ambiguity, false premises, sampling};
\node[cause=orange] (c5) at (11.3,2.5) {\textbf{Retrieval and context}\\ evidence missed, weak, or ignored};
\node[cause=orange] (c6) at (11.3,0.4) {\textbf{Multimodal grounding}\\ language priors beat visual evidence};
\foreach \i/\n in {c1/1,c2/2,c3/3,c4/4,c5/5,c6/6}
  \node[badge] at (\i.north west) {\n};
\node[hub]  (m) at (6.9,2.5)  {\textbf{Trained LLM}\\[2pt] knowledge frozen in its weights};
\node[outn] (o) at (16.1,2.5) {\textbf{Hallucination}\\[2pt] fluent but unsupported output};
\draw[arr] (5.0,2.5) -- (m.west);
\draw[arr] (m.east) -- (8.8,2.5);
\draw[arr] (13.8,2.5) -- (o.west);
\node[rectangle, rounded corners=4pt, draw=teal!60!black, thick, fill=teal!12, align=center, text width=13.2cm, minimum height=0.9cm, font=\footnotesize] (amp) at (6.9,-1.8) {\textbf{Domain and task amplifiers:} healthcare, spatial planning, code generation, where the same error costs more};
\draw[-{Stealth[length=2.6mm]}, thick, dashed, teal!60!black] (amp.east) -| (o.south);
\node[font=\scriptsize\itshape, text=teal!50!black, rotate=90] at (16.45,0.0) {raise the stakes};
\end{tikzpicture}}
\caption{Where the causes of hallucination act. Causes 1--3 are fixed when the model is trained; causes 4--6 operate each time it generates an answer. Domain and task do not cause hallucination on their own but amplify its cost. Numbers correspond to the subsections below.}
\label{fig:causes_map}
\end{figure}

Hallucinations can arise at multiple stages of an LLM's pipeline (Figure~\ref{fig:causes_map}). During training, noisy or incomplete datasets, limitations in parametric knowledge, and optimization objectives that prioritize likely or preferred responses over verified truth can contribute to errors \cite{Huang2025Survey,Wu2025LLM}. During inference, ambiguous prompts and stochastic decoding may produce unstable responses. Retrieval systems may miss, misrank, or fail to use relevant evidence, while multimodal models may misinterpret visual inputs \cite{Amugongo2025Retrieval,Farquhar2024Detecting,Yin2024MLLMsurvey}. These interacting causes are examined in Section~\ref{sec:causes}.

\section{Causes of LLM Hallucinations}
\label{sec:causes}
Hallucinations arise from interacting factors across the training and inference stages of an LLM. These include training data, parametric knowledge, optimization and alignment objectives, prompting and decoding strategies, retrieval and context handling, and multimodal grounding. The following subsections examine these factors along the model pipeline.

\subsection{Data and Training Corpus}
Pretraining corpora, the very large text collections on which an LLM first learns language, contain noisy, duplicated, biased, outdated, or incorrect content, and models reproduce such patterns as plausible statements \cite{Huang2025Survey}. Some hallucinations are therefore \textit{systematic}: the model repeats the same error across runs because it is strongly represented in the training distribution, that is, the statistical makeup of the training data. This differs from confabulation, where the answer changes arbitrarily from one run to the next \cite{Farquhar2024Detecting}. Data composition also matters across languages: hallucination is more frequent in low-resource directions (language pairs with little training text) and when translating out of English, and some toxic hallucinations, offensive content that appears in the output without being in the source, trace back to the training data \cite{Guerreiro2023Hallucinations}. In multimodal LLMs (MLLMs), which accept images alongside text, strong \textit{language priors}, the statistical habits learned from text alone, such as which words usually follow which, can dominate the visual evidence \cite{Yin2024MLLMsurvey}.

\subsection{Knowledge Limitations}
\textit{Parametric knowledge}, the facts stored implicitly in a model's weights rather than looked up from an external source, is static (frozen at training time) and incomplete\cite{Huang2025Survey,Amugongo2025Retrieval}. In code generation, models invent APIs (the functions and libraries a program can call), libraries, or functions that look reasonable but do not exist \cite{Liu2026Beyond}. Retrieval-augmented generation (RAG), in which relevant documents are fetched at query time and supplied to the model, reduces this dependence on static knowledge but introduces its own failures (Section~\ref{subsec:retrieval-causes}) \cite{Amugongo2025Retrieval,Miao2024Integrating}.

\subsection{Optimization, Fine-Tuning, and Alignment}
Pretraining optimizes next-token likelihood: the model is rewarded for predicting the most probable next word, not for checking whether the resulting statement is true. Instruction following (training the model to carry out user commands) and preference optimization (training on judgments of which answers are better, as in reinforcement learning from human feedback, RLHF) improve usefulness but create trade-offs among helpfulness, consistency, and truthfulness, and can favor producing an answer over expressing uncertainty \cite{Huang2025Survey}.Fine-tuning (additional training on a custom dataset relevant to your goals or task) is helpful only in the case of good-quality and diverse data available as a base. \cite{Wu2025LLM}.

\subsection{Prompting and Inference}
Prompt wording and inference settings can influence a model's response. Ambiguous instructions, false premises, missing context, or conflicting constraints may lead to unsupported answers. While structured and descriptive prompts can help reduce ambiguity in general, it is still impossible to be guaranteed by any chance of factual correctness. Adversarial prompts may also elicit hallucinations from otherwise capable models \cite{Omar2025MultiModel}. Some errors are seed-sensitive confabulations, meaning that the answer changes with the random seed used during sampling. Simply changing the temperature in isolation, which controls the randomness of token selection, may not be enough to completely eradicate them.\cite{Farquhar2024Detecting,Omar2025MultiModel}.

\subsection{Retrieval and Context}
\label{subsec:retrieval-causes}

RAG pushes some of the knowledge load outside into external proof while adding its own points of failure: the required documents may not be downloaded, any retrieved fragment may have little relevance to the demand, or RAG can access the evidence but fail to utilize it\cite{Amugongo2025Retrieval}. When it comes to lengthy contexts, the evidence might go unnoticed depending on where it is located, the 'lost-in-the-fullness' principle implying that the information presented in the middle of a long piece of text is less likely to be used than the evidence given at the beginning or at the end of it, which was proved by medical question-answering research\cite{Zhang2025Leveraging}. In other words, RAG enables a shift in the pathway to hallucination.

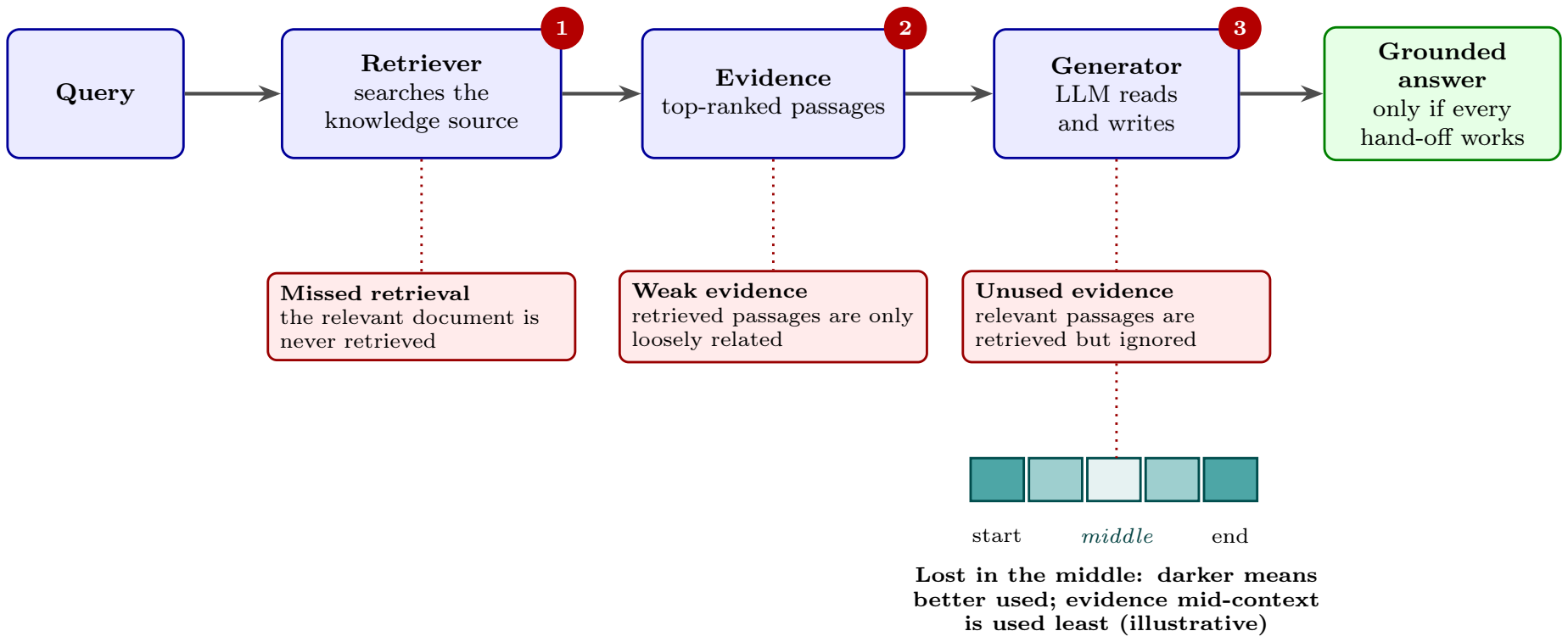
\begin{figure}[H]
\centering
\resizebox{\linewidth}{!}{%
\begin{tikzpicture}[
  stg/.style={rectangle, rounded corners=4pt, draw=blue!60!black, thick, fill=blue!8, align=center, font=\footnotesize, minimum height=1.5cm, inner sep=5pt},
  good/.style={stg, draw=green!50!black, fill=green!10},
  fail/.style={rectangle, rounded corners=3pt, draw=red!60!black, thick, fill=red!8, align=left, font=\scriptsize, text width=3.3cm, inner sep=4pt},
  fmark/.style={circle, fill=red!70!black, text=white, inner sep=1pt, minimum size=0.5cm, font=\scriptsize\bfseries},
  arr/.style={-{Stealth[length=2.6mm]}, very thick, black!70},
  dot/.style={dotted, thick, red!60!black}
]
\hyphenpenalty=10000 \exhyphenpenalty=10000
\node[stg, text width=1.7cm] (q) at (0,0) {\textbf{Query}};
\node[stg, text width=2.9cm] (r) at (3.8,0) {\textbf{Retriever}\\ searches the knowledge source};
\node[stg, text width=2.7cm] (p) at (7.9,0) {\textbf{Evidence}\\ top-ranked passages};
\node[stg, text width=2.5cm] (g) at (11.9,0) {\textbf{Generator}\\ LLM reads and writes};
\node[good, text width=2.4cm] (a) at (15.7,0) {\textbf{Grounded answer}\\ only if every hand-off works};
\draw[arr] (q)--(r); \draw[arr] (r)--(p); \draw[arr] (p)--(g); \draw[arr] (g)--(a);
\node[fmark] at (r.north east) {1};
\node[fmark] at (p.north east) {2};
\node[fmark] at (g.north east) {3};
\node[fail] (f1) at (3.8,-2.6)  {\textbf{Missed retrieval}\\ the relevant document is never retrieved};
\node[fail] (f2) at (7.9,-2.6)  {\textbf{Weak evidence}\\ retrieved passages are only loosely related};
\node[fail] (f3) at (11.9,-2.6) {\textbf{Unused evidence}\\ relevant passages are retrieved but ignored};
\draw[dot] (r.south) -- (f1.north);
\draw[dot] (p.south) -- (f2.north);
\draw[dot] (g.south) -- (f3.north);
\foreach \i/\c in {0/70,1/38,2/10,3/38,4/70}
  \draw[draw=teal!60!black, fill=teal!\c, thick] ({10.2+\i*0.68},-4.75) rectangle ++(0.62,0.5);
\node[font=\scriptsize] at (10.51,-5.15) {start};
\node[font=\scriptsize\itshape, text=teal!50!black] at (11.9,-5.15) {middle};
\node[font=\scriptsize] at (13.23,-5.15) {end};
\draw[dot] (f3.south) -- (11.9,-4.25);
\node[font=\scriptsize\bfseries, align=center, text width=5.2cm, anchor=north] at (11.9,-5.4) {Lost in the middle: darker means better used; evidence mid-context is used least (illustrative)};
\end{tikzpicture}}
\caption{Main points of failure regarding retrieval in RAG. A missing document, a document that comes up with not related passages or providing evidence that the generator fails to utilize in some way. The inset illustrates lost-in-the-middle effect, where evidence placed somewhere in the middle of a long context receives less attention }
\label{fig:rag_failures}
\end{figure}

\subsection{Multimodal Grounding and Domain Amplifiers}
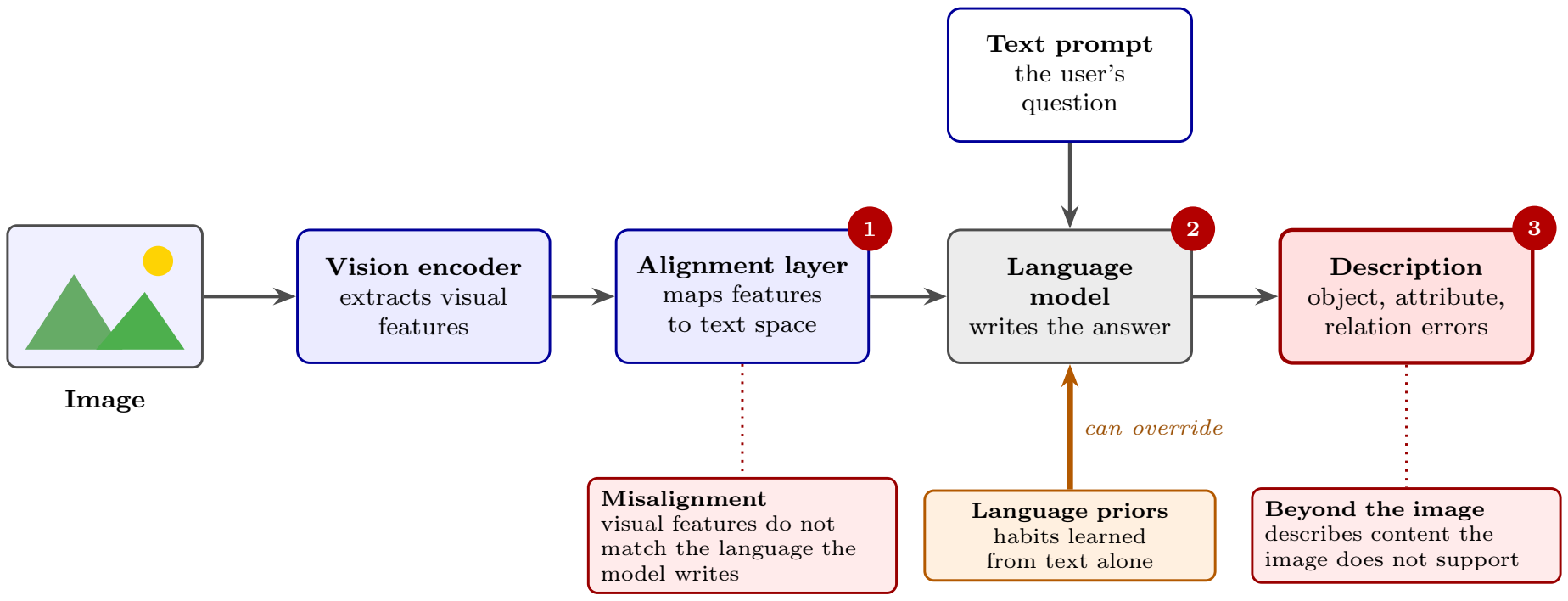
\begin{figure}[H]
\centering
\resizebox{\linewidth}{!}{%
\begin{tikzpicture}[
  stg/.style={rectangle, rounded corners=4pt, draw=blue!60!black, thick, fill=blue!8, align=center, font=\footnotesize, minimum height=1.5cm, inner sep=5pt},
  fail/.style={rectangle, rounded corners=3pt, draw=red!60!black, thick, fill=red!8, align=left, font=\scriptsize, text width=3.2cm, inner sep=4pt},
  fmark/.style={circle, fill=red!70!black, text=white, inner sep=1pt, minimum size=0.5cm, font=\scriptsize\bfseries},
  arr/.style={-{Stealth[length=2.6mm]}, very thick, black!70},
  dot/.style={dotted, thick, red!60!black}
]
\hyphenpenalty=10000 \exhyphenpenalty=10000
\node[rectangle, rounded corners=3pt, draw=black!70, thick, fill=blue!6, minimum width=2.2cm, minimum height=1.6cm] (img) at (0,0) {};
\begin{scope}[shift={(img.center)}]
  \fill[green!45!black!60] (-0.9,-0.6) -- (-0.35,0.25) -- (0.2,-0.6) -- cycle;
  \fill[green!55!black!70] (-0.1,-0.6) -- (0.45,0.05) -- (0.9,-0.6) -- cycle;
  \fill[yellow!70!orange] (0.6,0.4) circle (0.17);
\end{scope}
\node[font=\footnotesize\bfseries, below=0.12cm of img] {Image};
\node[stg, text width=2.5cm] (ve) at (3.6,0)  {\textbf{Vision encoder}\\ extracts visual features};
\node[stg, text width=2.5cm] (al) at (7.2,0)  {\textbf{Alignment layer}\\ maps features to text space};
\node[stg, text width=2.4cm, fill=gray!15, draw=black!70] (llm) at (10.9,0) {\textbf{Language model}\\ writes the answer};
\node[stg, text width=2.5cm, fill=red!12, draw=red!60!black, very thick] (out) at (14.7,0) {\textbf{Description}\\ object, attribute, relation errors};
\node[stg, text width=2.4cm, fill=white] (txt) at (10.9,2.5) {\textbf{Text prompt}\\ the user's question};
\draw[arr] (img.east)--(ve); \draw[arr] (ve)--(al); \draw[arr] (al)--(llm); \draw[arr] (llm)--(out);
\draw[arr] (txt)--(llm);
\node[fmark] at (al.north east)  {1};
\node[fmark] at (llm.north east) {2};
\node[fmark] at (out.north east) {3};
\node[rectangle, rounded corners=3pt, draw=orange!70!black, thick, fill=orange!12, align=center, font=\scriptsize, text width=3.0cm, inner sep=4pt] (lp) at (10.9,-2.7) {\textbf{Language priors}\\ habits learned from text alone};
\draw[-{Stealth[length=2.6mm]}, line width=2pt, orange!70!black] (lp.north) -- (llm.south) node[midway, right, font=\scriptsize\itshape, text=orange!60!black] {can override};
\node[fail] (f1) at (7.2,-2.7)  {\textbf{Misalignment}\\ visual features do not match the language the model writes};
\node[fail] (f3) at (14.7,-2.7) {\textbf{Beyond the image}\\ describes content the image does not support};
\draw[dot] (al.south) -- (f1.north);
\draw[dot] (out.south) -- (f3.north);
\end{tikzpicture}}
\caption{Multimodal grounding pathway in an LLM. Visual features are extracted by the vision encoder and aligned with the language model.}
\label{fig:mllm_grounding}
\end{figure}

The use of multimodal grounding brings new challenges as the large language models need to make the right connections between the visual space and the linguistic one. Errors could happen because the vision encoder does not recognize important nuances, the alignment layer fails to connect visual information accurately, or because linguistic assumptions override real visual data. Therefore, the models may invent objects, mistakenly identify certain characteristics, or state relationships not confirmed by the image. These mistakes can be especially critical in specific areas such as medicine, software engineering, and urban planning where some harmful or misleading decisions could be made. \cite{Yin2024MLLMsurvey,Yin2024Woodpecker}.  \cite{Asgari2025ClinicalSafety,Liu2026Beyond,Zhang2025Mitigating}. Figure~\ref{fig:mllm_grounding} illustrates this multimodal grounding process.

The factors described above indicate that hallucinations arise from interactions rather than the failure of one component. In our opinion, there are three important points to make. First, there is a difference in the nature of the various causes: some of them are established during training (such as data and knowledge) and lead to systematic errors, while others arise during inference (such as prompting, sampling, and retrieval) and produce varying confabulations. This distinction is critical from the viewpoint of taxonomy\cite{Farquhar2024Detecting}, since the signal available for one cause (e.g. disagreement among samples) cannot be applied to the other. Secondly, we view the training objective as the reason that is easy to overlook  \cite{Huang2025Survey}. A model that is rewarded for likelihood and helpfulness has little incentive to express its ignorance, so its overconfidence is a natural outcome of the model optimization process. Lastly, we should emphasize that retrieval and domain should be seen as amplifying the hallucination rather than eliminating it, because RAG simply moves the problem to the retrieval and application stages, while high-stakes domains increase the cost of the mistake but do not change its origin \cite{Amugongo2025Retrieval}. Since multiple causes can operate simultaneously, identifying a single culprit is unlikely to explain every hallucination.

\section{Detection of LLM Hallucinations}
\label{sec:detection}
Detection is the process that defines reliability of generation. It is believed that one needs reliable detection in order to assess these characteristics of LLM outputs effectively. Detection task is now becoming more difficult due to the advancement of language models: present hallucinations can be advanced and grammatically correct, thus making them credible and hard to find out. Detection is a key process in the\textit{generate--detect--verify--revise} correction pipeline. Here detector reveals suspicious output, then if it is confirmed, such output is altered or the output is rerouted to a different process (see figure below ).

\begin{figure}[H]
\centering
\resizebox{\linewidth}{!}{%
\begin{tikzpicture}[
  cell/.style={rectangle, rounded corners=3pt, align=center, font=\footnotesize,
               inner sep=3pt, minimum height=1.2cm, thick},
  fam/.style={cell, text width=3.2cm, draw=#1!60!black, fill=#1!25},
  qn/.style={cell,  text width=4.6cm, draw=#1!50!black, fill=#1!8},
  ev/.style={cell,  text width=4.4cm, draw=#1!50!black, fill=#1!8},
  vd/.style={cell,  text width=3.8cm, draw=#1!60!black, fill=#1!16},
  hdr/.style={rectangle, rounded corners=3pt, fill=gray!70!black, text=white,
              font=\footnotesize\bfseries, align=center, minimum height=0.7cm,
              inner sep=3pt},
  grpf/.style={rectangle, rounded corners=5pt, draw=#1!50, fill=#1!5,
               inner sep=6pt, line width=0.7pt},
  glab/.style={rotate=90, font=\scriptsize\bfseries, text=#1!60!black},
  arr/.style={-{Stealth[length=2mm]}, thick, black!60}
]
\hyphenpenalty=10000 \exhyphenpenalty=10000

\node[hdr, text width=3.2cm] at (0,1.55)     {Detection family};
\node[hdr, text width=4.6cm] at (4.45,1.55)  {Core question};
\node[hdr, text width=4.4cm] at (9.5,1.55)   {Evidence it relies on};
\node[hdr, text width=3.8cm] at (14.15,1.55) {Typical verdict};

\node[fam=blue] (F1) at (0,0) {\textbf{Human evaluation}\\[-1pt]{\scriptsize\cite{Ji2023Survey,Lin2024Towards}}};
\node[qn=blue]  (Q1) at (4.45,0)  {Is each statement supported by the source or the facts?};
\node[ev=blue]  (E1) at (9.5,0)   {Trained annotators};
\node[vd=blue]  (V1) at (14.15,0) {Scores, or labels per statement};

\node[fam=blue] (F2) at (0,-1.35) {\textbf{Reference-based (NLI)}\\[-1pt]{\scriptsize\cite{Huang2025Survey,Lin2024Towards}}};
\node[qn=blue]  (Q2) at (4.45,-1.35)  {Is the output consistent with a trusted source?};
\node[ev=blue]  (E2) at (9.5,-1.35)   {Source document, reference answer, or retrieved document};
\node[vd=blue]  (V2) at (14.15,-1.35) {Entailed, contradicted, or unsupported};

\node[fam=blue] (F3) at (0,-2.7) {\textbf{Retrieval, claim-level}\\[-1pt]{\scriptsize\cite{Huang2025Survey,Ji2023Survey}}};
\node[qn=blue]  (Q3) at (4.45,-2.7)  {Is each claim backed by external evidence?};
\node[ev=blue]  (E3) at (9.5,-2.7)   {Source documents, knowledge bases, search engines};
\node[vd=blue]  (V3) at (14.15,-2.7) {Supported, contradicted, or unresolved per claim};

\node[fam=orange] (F4) at (0,-4.55) {\textbf{Uncertainty}\\[-1pt]{\scriptsize\cite{Huang2025Survey,Shorinwa2025UQ,Farquhar2024Detecting}}};
\node[qn=orange]  (Q4) at (4.45,-4.55)  {Is the model unsure of what it generates?};
\node[ev=orange]  (E4) at (9.5,-4.55)   {Token probabilities, entropy, semantic entropy};
\node[vd=orange]  (V4) at (14.15,-4.55) {Uncertainty score};

\node[fam=orange] (F5) at (0,-5.9) {\textbf{Self-consistency}\\[-1pt]{\scriptsize\cite{Huang2025Survey,Ji2023Survey}}};
\node[qn=orange]  (Q5) at (4.45,-5.9)  {Do claims stay stable across samples?};
\node[ev=orange]  (E5) at (9.5,-5.9)   {Repeated generations of the same model};
\node[vd=orange]  (V5) at (14.15,-5.9) {Stable or unstable claims};

\node[fam=violet] (F6) at (0,-7.75) {\textbf{LLM-as-a-Judge}\\[-1pt]{\scriptsize\cite{AnhHoang2025Survey,Huang2025Survey}}};
\node[qn=violet]  (Q6) at (4.45,-7.75)  {Does a second model judge the response reliable?};
\node[ev=violet]  (E6) at (9.5,-7.75)   {Instructions, references, or evidence given to the judge};
\node[vd=violet]  (V6) at (14.15,-7.75) {Judgments or scores};

\node[fam=teal] (F7) at (0,-9.6) {\textbf{Multimodal}\\[-1pt]{\scriptsize\cite{Yin2024Woodpecker}}};
\node[qn=teal]  (Q7) at (4.45,-9.6)  {Does the text match the image?};
\node[ev=teal]  (E7) at (9.5,-9.6)   {Visual evidence};
\node[vd=teal]  (V7) at (14.15,-9.6) {Per-concept verification};

\node[fam=teal] (F8) at (0,-10.95) {\textbf{Multilingual}\\[-1pt]{\scriptsize\cite{Guerreiro2023Hallucinations}}};
\node[qn=teal]  (Q8) at (4.45,-10.95)  {Is the output detached from, or repeating, the source?};
\node[ev=teal]  (E8) at (9.5,-10.95)   {Source-contribution and n-gram signals};
\node[vd=teal]  (V8) at (14.15,-10.95) {Detached or oscillatory hallucination};

\foreach \i in {1,...,8}{
  \draw[arr] (F\i.east) -- (Q\i.west);
  \draw[arr] (Q\i.east) -- (E\i.west);
  \draw[arr] (E\i.east) -- (V\i.west);
}

\begin{scope}[on background layer]
  \node[grpf=blue,   fit=(F1)(F3)(V1)(V3)] {};
  \node[grpf=orange, fit=(F4)(F5)(V4)(V5)] {};
  \node[grpf=violet, fit=(F6)(V6)] {};
  \node[grpf=teal,   fit=(F7)(F8)(V7)(V8)] {};
\end{scope}
\node[glab=blue]   at (-2.15,-1.35)  {With trusted evidence};
\node[glab=orange] at (-2.15,-5.225) {Reference-free};
\node[glab=violet] at (-2.15,-7.75)  {Model-based};
\node[glab=teal]   at (-2.15,-10.275){Specialized settings};

\node[cell, text width=16.4cm, draw=gray!70, fill=gray!12, dashed] at (7.0,-12.7)
  {\textbf{No single family suffices:} practical systems combine several of them into hybrid pipelines.};
\end{tikzpicture}}
\caption{Detection families simply explained. An image shows that each row contains an overview of the question asked by a family, the reasoning its used and the conclusion it reached. The colors point how different detection families exclusively rely on the evidence presented.}
\label{fig:det-glance}
\end{figure}
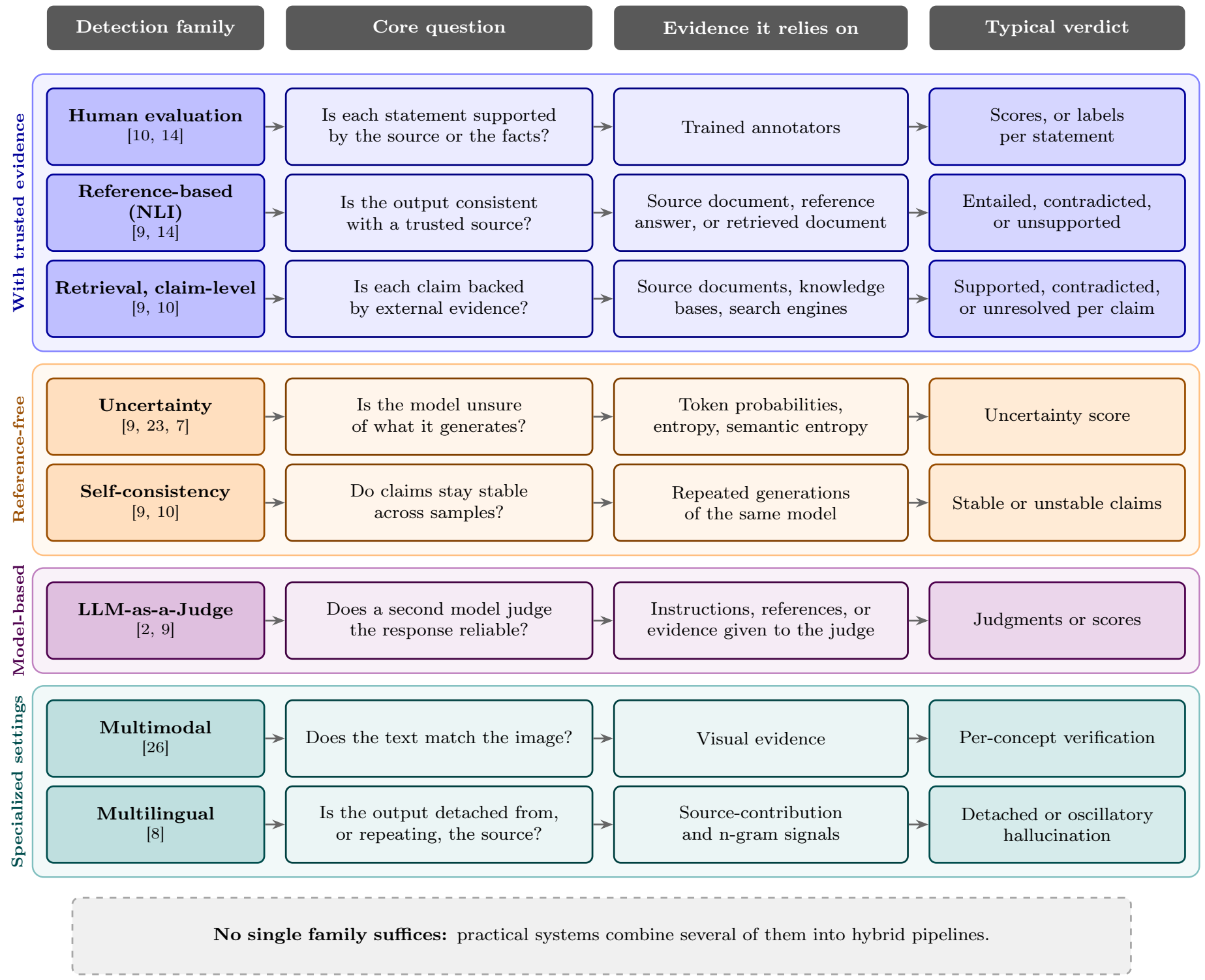

In annotators, statements can be categorized into three options: supported, contradicted, and unsupported. In entailment checks, statements are classified based on whether they are supported by, contradicted by, or unsupported by their source. In terms of claim-level pipelines, the outcomes can be summarized as either supported, contradicted, or unresolved\cite{Ji2023Survey,Huang2025Survey}. Depending on the type of task, the detection of such phenomena can differ\cite{Farquhar2024Detecting}. In particular, the system can encounter either intrinsic or extrinsic hallucination events during source-oriented tasks or similar issues when working on different types of content, including images or translation output \cite{Guerreiro2023Hallucinations}.

The above-mentioned mechanisms are based on the idea of being complementary instead of exclusive. As for the practical implementation of such methods, many systems utilize several techniques. The figure and table illustrate the importance of the evidence being available for the selection of the method. In addition, the content of this chapter will be discussed from the perspective of SolvedDetects methods.

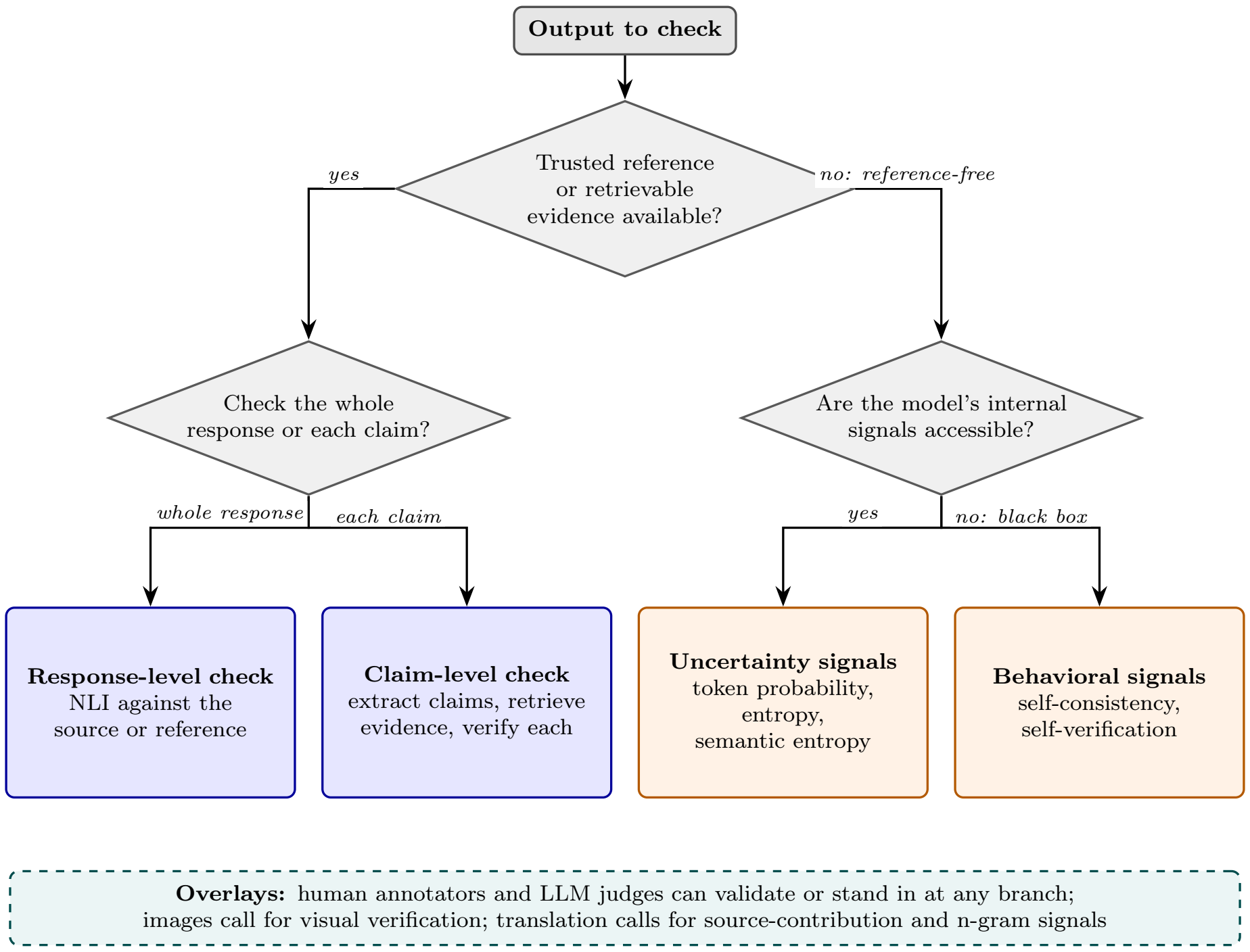
\begin{figure}[H]
\centering
\resizebox{\linewidth}{!}{%
\begin{tikzpicture}[
  n/.style={rectangle, rounded corners=3pt, draw, thick, align=center, font=\footnotesize, inner sep=5pt},
  q/.style={diamond, aspect=2.6, draw=gray!70!black, thick, fill=gray!12, align=center, font=\footnotesize, inner sep=1pt, text width=3.3cm},
  res/.style={n, text width=3.3cm, minimum height=2.4cm},
  arr/.style={-{Stealth[length=2.6mm]}, thick},
  lab/.style={font=\scriptsize\itshape, fill=white, inner sep=1.5pt}
]
\hyphenpenalty=10000 \exhyphenpenalty=10000
\node[n, fill=gray!20, draw=black!70] (s) at (7.9,0) {\textbf{Output to check}};
\node[q] (q1) at (7.9,-2.0) {Trusted reference or retrievable evidence available?};
\node[q] (q2) at (3.9,-4.9) {Check the whole response or each claim?};
\node[q] (q3) at (11.9,-4.9) {Are the model's internal signals accessible?};
\node[res, draw=blue!60!black, fill=blue!10]     (r1) at (1.9,-8.5)  {\textbf{Response-level check}\\ NLI against the source or reference};
\node[res, draw=blue!60!black, fill=blue!10]     (r2) at (5.9,-8.5)  {\textbf{Claim-level check}\\ extract claims, retrieve evidence, verify each};
\node[res, draw=orange!70!black, fill=orange!10] (r3) at (9.9,-8.5)  {\textbf{Uncertainty signals}\\ token probability, entropy, semantic entropy};
\node[res, draw=orange!70!black, fill=orange!10] (r4) at (13.9,-8.5) {\textbf{Behavioral signals}\\ self-consistency, self-verification};
\draw[arr] (s)--(q1);
\draw[arr] (q1.west) -| node[pos=0.3, above, lab] {yes} (q2.north);
\draw[arr] (q1.east) -| node[pos=0.3, above, lab] {no: reference-free} (q3.north);
\draw[arr] (q2.south) -- ++(0,-0.4) -| node[pos=0.25, above, lab] {whole response} (r1.north);
\draw[arr] (q2.south) -- ++(0,-0.4) -| node[pos=0.25, above, lab] {each claim} (r2.north);
\draw[arr] (q3.south) -- ++(0,-0.4) -| node[pos=0.25, above, lab] {yes} (r3.north);
\draw[arr] (q3.south) -- ++(0,-0.4) -| node[pos=0.25, above, lab] {no: black box} (r4.north);
\node[n, dashed, fill=teal!8, draw=teal!60!black, text width=15.2cm] (ov) at (7.9,-11.1) {\textbf{Overlays:} human annotators and LLM judges can validate or stand in at any branch; images call for visual verification; translation calls for source-contribution and n-gram signals};
\end{tikzpicture}}
\caption{Choosing a detection strategy. Whether trusted evidence is available, at what granularity the answer must be checked, and whether the model's internals are accessible together determine which family of detectors applies.}
\label{fig:det-choose}
\end{figure}

\subsection{Judging Against Evidence: Humans, References, and Inference}

\textbf{Human evaluation.} Human annotators provide an important reference standard for assessing factuality, faithfulness, and contextual consistency. They have the ability to determine factuality, fidelity, context consistency as well as complex semantic errors that can be missed by any single automated metric. They can assign factuality and hallucination scores to a response, compare it to reference responses, and qualify each statement stating whether it is supported, contradicted, or unsupported. In the case of source-based tasks, such as summarizing, annotation systems allow to distinguish intrinsic hallucinations, which are contradictory to the source of material, and extrinsic ones that bring unsupported information, and they allow to measure even more general features, such as accuracy and informativeness, when some answer mixes unsupported additions and lacks some content. The main limitation is scalability, since annotating is very labor-intensive and slow when it comes to big benchmarks, and the more narrow is the field of study, the more complicated is the task of manual evaluation.

\textbf{Reference-based detection.} In reference-based detection, the response is compared to credible evidence, whether it's a reference answer, retrieved document, or any other form of verified proof. With evaluation not derived from the behavior of the model, the method is appropriate for grounded generation tasks. More traditional metrics of overlap, BLEU, ROUGE, and METEOR are merely surface indicators, by themselves ineffective for determining facts and evidence; the progress of research led to the usage of factuality-oriented techniques instead. The most widely used approach is natural language inference (NLI), which determines whether a statement is entailed by, contradicted by, or unsupported by the available evidence. The validity of the conclusions depends on the evidence; the results are only as strong as the quality and reliability of the reference.

Automated checks can be classified into five groups.  Table~\ref{tab:det-families} summarizes them and shows their location in this section.

\begin{table}[H]
\centering
\small
\renewcommand{\arraystretch}{1.25}
\caption{Five families of automated hallucination-detection methods.}
\label{tab:det-families}
\begin{tabular}{|L{0.17\linewidth}|L{0.50\linewidth}|L{0.19\linewidth}|}
\hline
\rowcolor{gray!15}
\textbf{Family} & \textbf{How it checks the output} \\
\hline
Fact-based & Compares the important information in the source with that in the generated output\\
\hline
Classifier-based & Uses entailment or contradiction models to decide whether statements are supported by the source  \\
\hline
QA-based & Generates questions from the source and checks whether the generated output answers them consistently\\
\hline
Uncertainty-based & Estimates the likelihood that a generation is unreliable \cite{Farquhar2024Detecting}  \\
\hline
LLM-based & Prompts another language model to evaluate the quality or factuality of the response \\
\hline
\end{tabular}
\end{table}

\subsection{Verifying Claim by Claim}
\label{subsec:det-claims}

\begin{figure}[H]
\centering
\resizebox{\linewidth}{!}{%
\begin{tikzpicture}[
  stg/.style={rectangle, rounded corners=4pt, draw=blue!60!black, thick, fill=blue!8, align=center, font=\footnotesize, inner sep=5pt, minimum height=1.6cm},
  cl/.style={rectangle, rounded corners=3pt, draw=black!60, thick, fill=white, align=center, font=\scriptsize, minimum width=1.7cm, minimum height=0.7cm},
  vd/.style={rectangle, rounded corners=3pt, thick, align=center, font=\scriptsize, text width=2.3cm, minimum height=0.9cm},
  fail/.style={rectangle, rounded corners=3pt, draw=red!60!black, thick, fill=red!8, align=left, font=\scriptsize, text width=4.6cm, inner sep=4pt},
  arr/.style={-{Stealth[length=2.4mm]}, thick, black!70}
]
\hyphenpenalty=10000 \exhyphenpenalty=10000
\node[stg, text width=2.0cm, fill=gray!15, draw=black!70] (resp) at (0,0) {\textbf{Generated response}\\ one fluent paragraph};
\node[stg, text width=2.2cm] (ext) at (3.6,0) {\textbf{1. Claim extraction}\\ split into checkable statements};
\node[cl] (c1) at (6.7,1.3)  {claim 1};
\node[cl] (c2) at (6.7,0)    {claim 2};
\node[cl] (c3) at (6.7,-1.3) {claim 3};
\node[stg, text width=2.2cm] (ret) at (9.9,0) {\textbf{2. Evidence retrieval}\\ documents, knowledge bases, search};
\node[stg, text width=2.2cm] (ver) at (13.2,0) {\textbf{3. Claim verification}\\ compare each claim with its evidence};
\node[vd, draw=green!50!black, fill=green!12]  (v1) at (16.5,1.3)  {supported};
\node[vd, draw=red!60!black,   fill=red!12]    (v2) at (16.5,0)    {contradicted};
\node[vd, draw=orange!70!black, fill=orange!14] (v3) at (16.5,-1.3) {unresolved};
\draw[arr] (resp)--(ext);
\foreach \c in {c1,c2,c3}{\draw[arr] (ext.east)--(\c.west); \draw[arr] (\c.east)--(ret.west);}
\draw[arr] (ret)--(ver);
\draw[arr] (ver.east)--(v1.west); \draw[arr] (ver.east)--(v2.west); \draw[arr] (ver.east)--(v3.west);
\node[font=\scriptsize\itshape, text=black!70] at (6.7,-2.15) {illustrative claims};
\node[fail] (f1) at (10.9,-3.3) {\textbf{Weak point:} missing, outdated, noisy, or irrelevant evidence can give a wrong verdict even when the claim is correct};
\draw[dotted, thick, red!60!black] (ret.south) -- (ret.south |- f1.north);
\node[rectangle, rounded corners=3pt, draw=teal!60!black, thick, fill=teal!8, align=left, font=\scriptsize, text width=4.6cm, inner sep=4pt] (g) at (4.4,-3.3) {\textbf{Gain:} a response-level label would hide that only some claims fail; here the failing statement is located};
\end{tikzpicture}}
\caption{Claim-level detection. The response is split into independently checkable claims, each claim is matched with retrieved evidence, and each receives its own verdict. The three claims are illustrative.}
\label{fig:det-claims}
\end{figure}
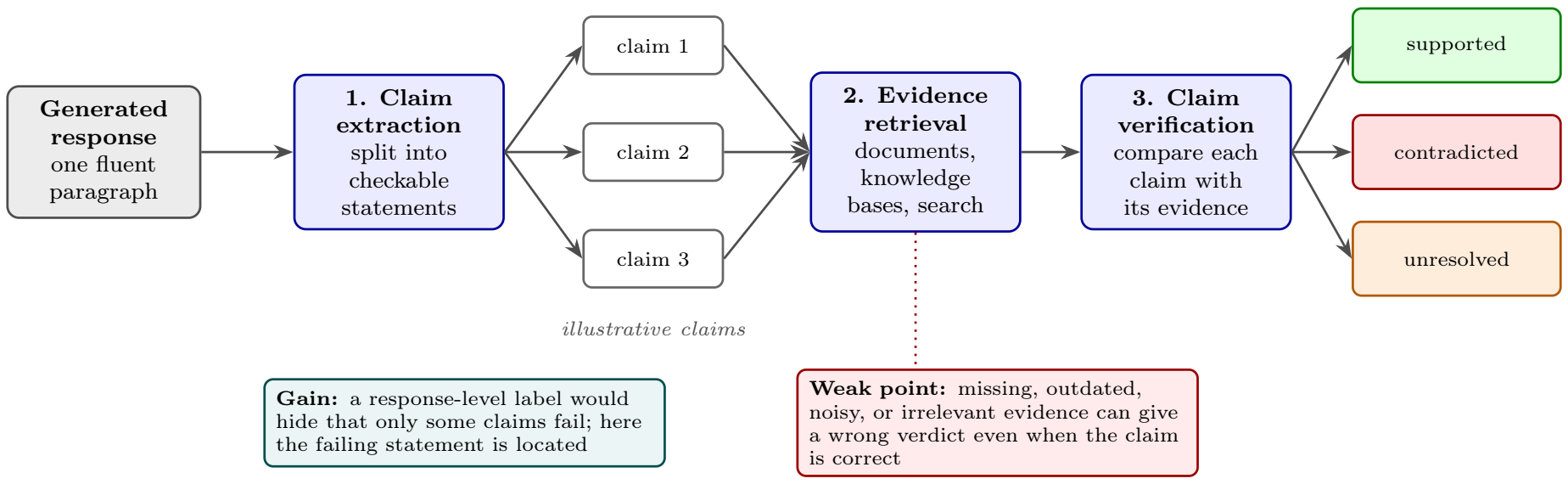

A response is seldom entirely correct or incorrect. A response may contain a mixture of correct and incorrect claims, which a single response-level label may fail to capture. So Claim-level detection splits the response into smaller components, thus allowing for their independent verification.
 Evidence can consist of documents, databases or repositories of data, and each claim is verified as being either supported, contradicted, or unresolved. This makes the process applicable to such areas as open-domain question answering and retrieval-augmented generation (RAG) where the verification process can be linked to the evidence utilized for answering questions. Figure~\ref{fig:det-claims} illustrates the claim-level verification workflow.

Claim-level verification carries over the problems stemming from retrieval. Lacking, outdated, noisy, or irrelevant evidence can make the verification result incorrect. Furthermore, a wrong final answer does not necessarily tell us if the error occurred in retrieval or generation. Breaking an answer down into claims can point to unsupported claims but it does not explain whether the reasoning itself was correct. Thus, detectors should evaluate reasoning consistency along with the correctness of the final answer.~\cite{Li2024Drowzee}.

\subsection{Judging Without Evidence: Uncertainty and Consistency}
\label{subsec:det-free}
When we don’t have a valid reference or the original, our reference-free approach assesses whether the model is exhibiting uncertainty, inconsistency, or various kinds of unreliability.~\cite{Huang2025Survey,Ji2023Survey,Shorinwa2025UQ}.

\textbf{Token probability and entropy} Internal signals are the easiest to identify: low token probability or high predictive entropy indicates that the model has low confidence in the output\cite{Huang2025Survey,Shorinwa2025UQ}. This is useful information because it reveals something about the model itself, which is valuable when external confirmation is impossible or costly. However, token uncertainty is not the same thing as factual uncertainty: it is possible to state something true using low-probability tokens, while a very probable result may still not be true.

\textbf{Semantic entropy} Farquhar argues that the standard sequence uncertainty could be deceiving since various sequences of words convey the same meaning. Semantic entropy thus classifies generations through semantic equivalence while measuring uncertainty based on meanings, not sequences of tokens. The method was invented for the identification of confabulations and random and unsupported fact generators, but it requires no task-specific training labels. Farquhar et al.\ evaluate semantic entropy using the area under the receiver operating characteristic curve (AUROC) and the area under the rejection-accuracy curve (AURAC), which measures the benefit of refusing answers judged likely to produce confabulations~\cite{Farquhar2024Detecting}. Hence, the method is favorable for scalable internal detection, making it a useful reference-free signal for detecting confabulations~\cite{Farquhar2024Detecting}.

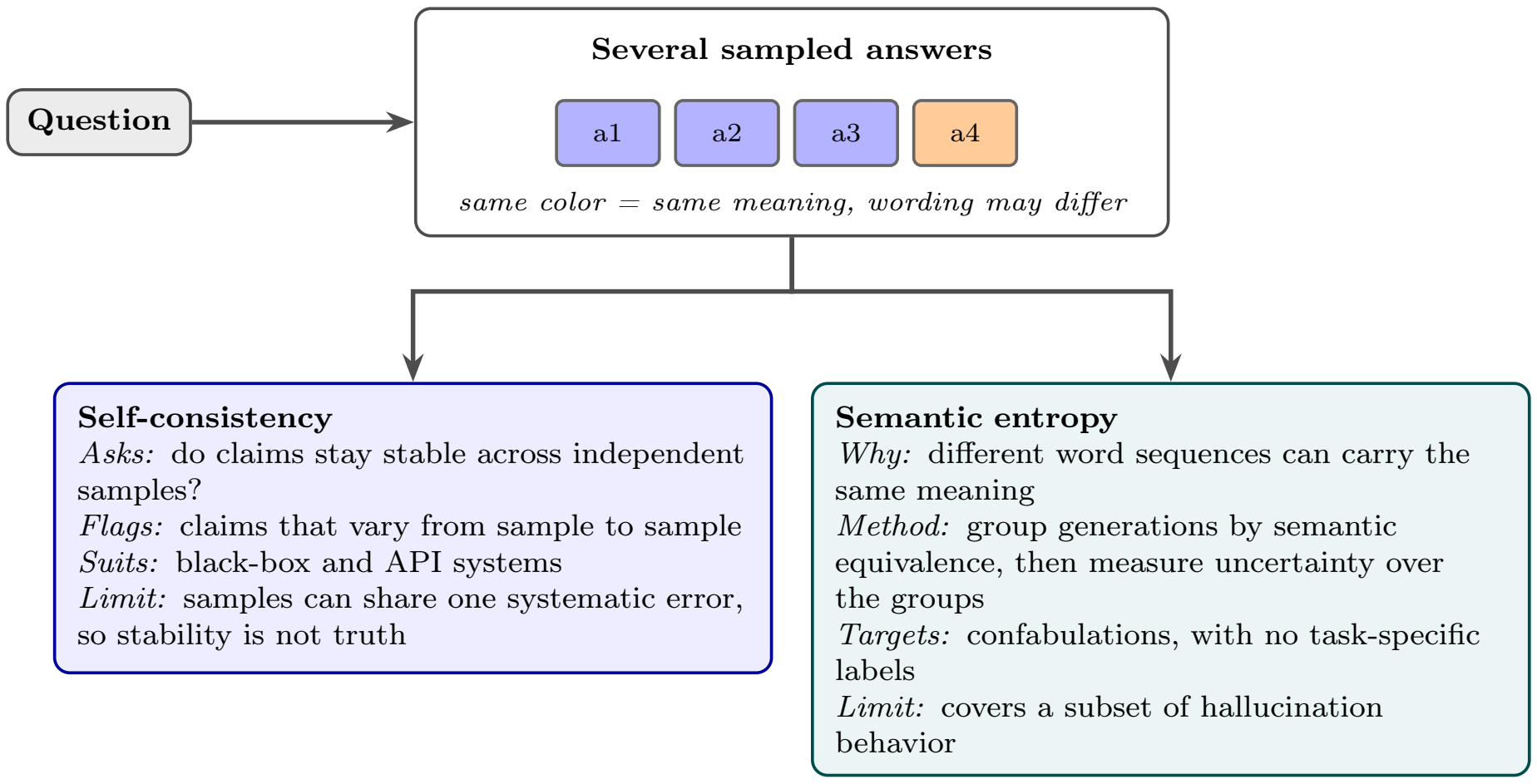
\begin{figure}[H]
\centering
\resizebox{\linewidth}{!}{%
\begin{tikzpicture}[
  n/.style={rectangle, rounded corners=4pt, draw, thick, align=center, font=\footnotesize, inner sep=5pt},
  chip/.style={rectangle, rounded corners=2pt, draw=black!60, thick, minimum width=0.95cm, minimum height=0.6cm, font=\scriptsize},
  box/.style={rectangle, rounded corners=4pt, thick, align=left, font=\footnotesize, text width=6.2cm, inner sep=6pt},
  arr/.style={-{Stealth[length=2.6mm]}, very thick, black!70}
]
\hyphenpenalty=10000 \exhyphenpenalty=10000
\node[n, fill=gray!15, draw=black!70] (qn) at (0,0) {\textbf{Question}};
\node[n, fill=white, draw=black!70, text width=6.6cm, minimum height=2.1cm] (sm) at (6.4,0) {};
\node[font=\footnotesize\bfseries] at (6.4,0.65) {Several sampled answers};
\node[chip, fill=blue!30]   at (4.7,-0.1) {a1};
\node[chip, fill=blue!30]   at (5.8,-0.1) {a2};
\node[chip, fill=blue!30]   at (6.9,-0.1) {a3};
\node[chip, fill=orange!40] at (8.0,-0.1) {a4};
\node[font=\scriptsize\itshape] at (6.4,-0.75) {same color = same meaning, wording may differ};
\draw[arr] (qn)--(sm);
\node[box, draw=blue!60!black, fill=blue!7, anchor=north] (sc) at (2.9,-2.4) {\textbf{Self-consistency}\\ \textit{Asks:} do claims stay stable across independent samples?\\ \textit{Flags:} claims that vary from sample to sample\\ \textit{Suits:} black-box and API systems\\ \textit{Limit:} samples can share one systematic error, so stability is not truth};
\node[box, draw=teal!60!black, fill=teal!8, anchor=north] (se) at (9.9,-2.4) {\textbf{Semantic entropy}\\ \textit{Why:} different word sequences can carry the same meaning\\ \textit{Method:} group generations by semantic equivalence, then measure uncertainty over the groups\\ \textit{Targets:} confabulations, with no task-specific labels\\ \textit{Limit:} covers a subset of hallucination behavior};
\draw[arr] (sm.south) -- ++(0,-0.5) -| (sc.north);
\draw[arr] (sm.south) -- ++(0,-0.5) -| (se.north);
\end{tikzpicture}}
\caption{Various techniques have been developed for analyzing samples developed using a single model. Self-consistency checks whether the results being produced are consistent, while semantic entropy collects the results based on their meaning and evaluates uncertainty on such groups of results. The collected responses represent an example of this process.}
\label{fig:det-sampling}
\end{figure}

\textbf{Self-consistency and behavioral probes} The second category of signals looks at the operation of the model over multiple generations. Claims that are stable over the variability of the data are thought to be more accurate than claims that change over that data. Other methods make use of indirect questioning, self-verification, reconstruction, and repeated generation to find inconsistencies in the answers provided by the model. Such methods are appropriate for black-box models and APIs since they do not require access to the internal state of the model. Figure~\ref{fig:det-sampling} compares self-consistency and semantic entropy, highlighting the difference between measuring response stability and measuring uncertainty over semantic meanings.

\textbf{To what extent should we have faith in uncertainty?}Uncertainty indications must be seen as indirect indicators of risk, not as measures of truth. A model can give high confidence in wrong answers, as established in research on clinical reason. While semantic entropy is able to detect a group of hallucinations, particularly confabulations, it does not reveal every case of systematic errors. The same goes for repeated generations that may show the same systematic error again and again. Therefore, consistency indicates stability, not accuracy.~\cite{Kim2025ClinicalReasoning} ~\cite{Farquhar2024Detecting}

\begin{table}[H]
\centering
\small
\renewcommand{\arraystretch}{1.25}
\caption{Reference-free signals: what each measures, what it needs, and its main caveat.}
\label{tab:det-signals}
\begin{tabular}{|L{0.17\linewidth}|L{0.22\linewidth}|L{0.17\linewidth}|L{0.28\linewidth}|}
\hline
\rowcolor{gray!15}
\textbf{Signal} & \textbf{What it measures} & \textbf{What it needs} & \textbf{Main caveat} \\
\hline
Token probability, predictive entropy \cite{Huang2025Survey,Shorinwa2025UQ} & The model's own uncertainty about the tokens it generates & Signals taken directly from the model & Token-level uncertainty is not factual uncertainty \\
\hline
Semantic entropy \cite{Farquhar2024Detecting} & Uncertainty over meanings after grouping equivalent generations & Several generations; no task-specific labels & Targets a subset of hallucinations; may miss systematic errors \\
\hline
Self-consistency \cite{Huang2025Survey,Ji2023Survey} & Stability of claims across independently sampled responses & Repeated generations only & Systematic errors can stay consistent \\
\hline
Indirect questioning, self-verification, reconstruction \cite{Huang2025Survey} & Inconsistencies exposed in the model's own answers & Further queries to the same model, no internal access needed & Consistency indicates stability, not truth \\
\hline
Calibration view of uncertainty \cite{Shorinwa2025UQ,Kim2025ClinicalReasoning} & Whether confidence matches correctness & Confidence estimates alongside correctness & Confident models can still hallucinate; overconfidence reported in clinical reasoning \\
\hline
\hline
\end{tabular}
\end{table}

\subsection{Letting a Model Judge}
\label{subsec:det-judge}

With the increase in their capabilities, the use of LLMs in determining truthfulness of production from other models has also increased. In {LLM-as-a-Judge} schemes, some form of language model is utilized to compare the output with the specific instructions or guidelines, the source of evidence, or any evaluation criteria value . LLM methods can be helpful in scaling evaluation and in identifying semantic differences that may not be found using other metrics, but at the same time, their evaluations must be measured against human annotators. 

Nonetheless, LLM approaches can themselves be flawed, unreliable, and inconsistent. This is the reason they need to be contrasted with human-produced annotation and also tested on different tasks and model families~\cite{Li2024Drowzee}. In multimodal tasks, judges must also be provided with visual materials since there are no reliable ways for textual judges to correct for errors that may arise due to a visual aspect in evaluating the audiovisual information that contains a visuals-based piece of input or process.~\cite{Yin2024MLLMsurvey,Yin2024Woodpecker,Chen2025MMHallucination}.

\subsection{Beyond English Text: Images, Languages, and New Domains}

\paragraph{Multimodal detection} For image-grounded processes, detectors must make comparisons between the descriptions produced and the evidence from the image and language criteria. Some of the possible mistakes are made by hallucinatory objects, wrong properties, and misleading objects' connections. Woodpecker suggests using a systematic way of processing the information which allows recognizing important visual ideas. The system creates the questions related to them, checks whether the data received from the image is correct, makes visual claims, and improves the faulty claims. The experiment made with the help of POPE benchmark proves that structured visual verification is very effective. Therefore, multimodal detection requires visual grounding along with evaluation in language~\cite{Yin2024Woodpecker,Yin2024MLLMsurvey,Chen2025MMHallucination}.

\paragraph{Multilingual detection} Guerreiro et al.\ analyzed more than 100 translation directions and found that behavior differs substantially across resource levels and directions \cite{Guerreiro2023Hallucinations}. They distinguish \textit{detached} hallucinations, where the output is weakly grounded in the source, from repetitive, \textit{oscillatory} behavior, and they studied source-contribution signals such as ALTI+ for the former and n-gram-based signals for the latter \cite{Guerreiro2023Hallucinations}. Hallucinations are especially prevalent in low-resource directions and when translating out of English, and detectors developed for English monolingual generation may not transfer \cite{Guerreiro2023Hallucinations}.

\paragraph{Domain and task shift} A detector that performs well on open-domain factual question answering may not do so on clinical reasoning, code generation, summarization, or long-context generation \cite{Farquhar2024Detecting,Huang2025Survey}. Robust evaluation of a detector should therefore include unseen domains and task shifts rather than rely on in-domain tests alone. Table~\ref{tab:det-special} summarizes the major detection challenges and approaches in specialized settings, including multimodal, multilingual, domain-specific, and reasoning-dependent tasks.

\begin{table}[H]
\centering
\small
\renewcommand{\arraystretch}{1.25}
\caption{Detection in specialized settings.}
\label{tab:det-special}
\begin{tabular}{|L{0.14\linewidth}|L{0.22\linewidth}|L{0.24\linewidth}|L{0.25\linewidth}|}
\hline
\rowcolor{gray!15}
\textbf{Setting} & \textbf{What goes wrong} & \textbf{Method or signal} & \textbf{What the evidence shows} \\
\hline
Multimodal \cite{Yin2024Woodpecker,Yin2024MLLMsurvey,Chen2025MMHallucination} & Wrong object existence, attributes, or relationships & Woodpecker: key concepts, questions, visual validation, visual claims, correction & Improvement reported on POPE; text-only metrics are inadequate on their own \\
\hline
Multilingual translation \cite{Guerreiro2023Hallucinations} & Detached output, or repetitive oscillation & ALTI+ source contribution; n-gram signals & Over 100 directions studied; low-resource and out-of-English cases worse; English detectors may not transfer \\
\hline
Domain and task shift \cite{Farquhar2024Detecting,Huang2025Survey} & A detector tuned on one task fails on others & Evaluation on unseen domains and task shifts & Open-domain QA performance may not carry over to clinical, code, summarization, or long-context tasks \\
\hline
Reasoning-dependent answers \cite{Li2024Drowzee} & Conclusions rest on complex logical relations & Claim-level checks & Do not fully resolve reasoning-level failures; an answer can be right despite wrong reasoning \\
\hline
\end{tabular}
\end{table}

\subsection{Putting Detectors Together}
\label{subsec:det-combine}

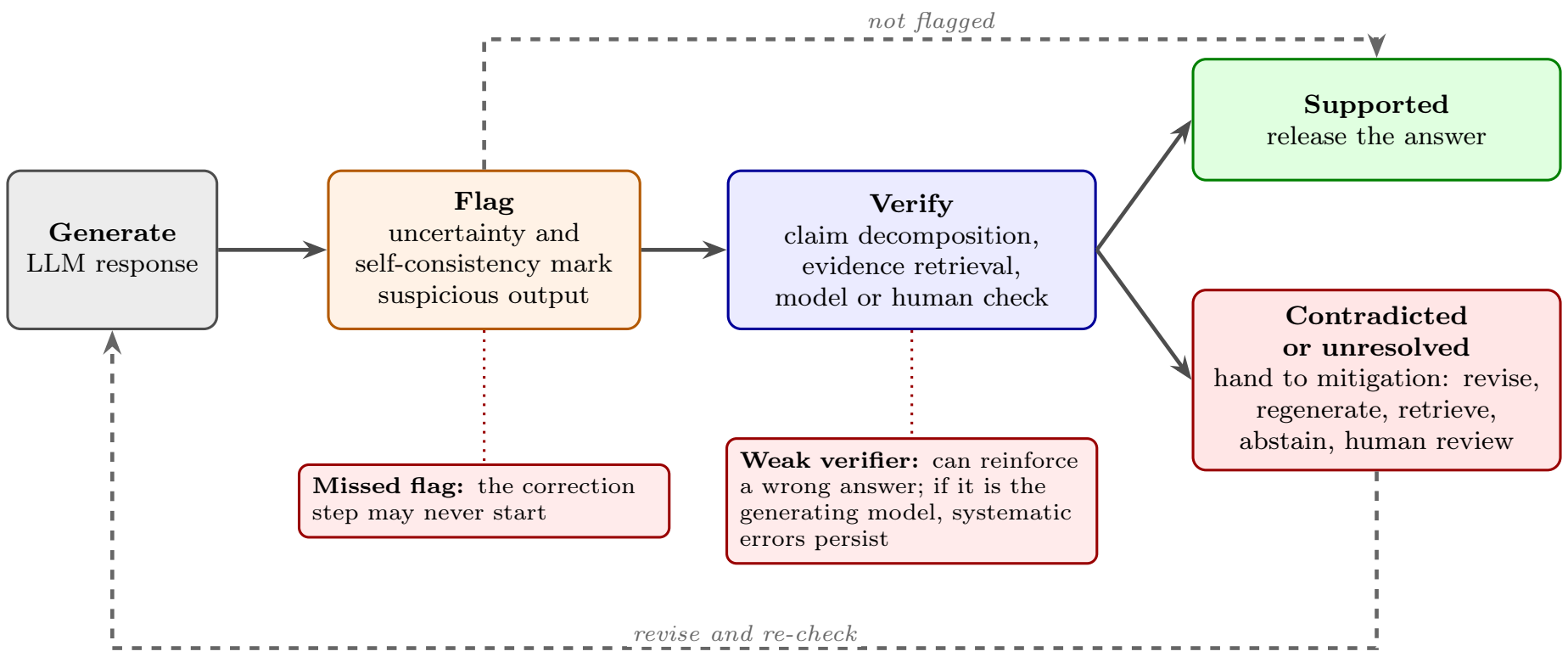
\begin{figure}[H]
\centering
\resizebox{\linewidth}{!}{%
\begin{tikzpicture}[
  stg/.style={rectangle, rounded corners=4pt, draw=blue!60!black, thick, fill=blue!8, align=center, font=\footnotesize, inner sep=5pt, minimum height=1.7cm},
  oc/.style={rectangle, rounded corners=4pt, thick, align=center, font=\footnotesize, text width=3.6cm, inner sep=5pt, minimum height=1.3cm},
  fail/.style={rectangle, rounded corners=3pt, draw=red!60!black, thick, fill=red!8, align=left, font=\scriptsize, text width=3.7cm, inner sep=4pt},
  arr/.style={-{Stealth[length=2.6mm]}, very thick, black!70},
  lab/.style={font=\scriptsize\itshape, fill=white, inner sep=1.5pt}
]
\hyphenpenalty=10000 \exhyphenpenalty=10000
\node[stg, text width=1.9cm, fill=gray!15, draw=black!70] (gen) at (0,0) {\textbf{Generate}\\ LLM response};
\node[stg, text width=3.0cm, draw=orange!70!black, fill=orange!10] (flag) at (4.0,0) {\textbf{Flag}\\ uncertainty and self-consistency mark suspicious output};
\node[stg, text width=3.6cm] (ver) at (8.6,0) {\textbf{Verify}\\ claim decomposition, evidence retrieval, model or human check};
\node[oc, draw=green!50!black, fill=green!12] (rel) at (13.6,1.4) {\textbf{Supported}\\ release the answer};
\node[oc, draw=red!60!black, fill=red!10] (mit) at (13.6,-1.4) {\textbf{Contradicted or unresolved}\\ hand to mitigation: revise, regenerate, retrieve, abstain, human review};
\draw[arr] (gen)--(flag); \draw[arr] (flag)--(ver);
\draw[arr] (ver.east) -- (rel.west); \draw[arr] (ver.east) -- (mit.west);
\draw[arr, dashed, black!60] (flag.north) -- ++(0,1.4) -| (rel.north) node[pos=0.25, above, lab] {not flagged};
\draw[arr, dashed, black!60] (mit.south) -- ++(0,-1.9) -| (gen.south) node[pos=0.25, above, lab] {revise and re-check};
\node[fail] (f1) at (4.0,-2.7) {\textbf{Missed flag:} the correction step may never start};
\node[fail] (f2) at (8.6,-2.7) {\textbf{Weak verifier:} can reinforce a wrong answer; if it is the generating model, systematic errors persist};
\draw[dotted, thick, red!60!black] (flag.south) -- (f1.north);
\draw[dotted, thick, red!60!black] (ver.south) -- (f2.north);
\end{tikzpicture}}
\caption{A hybrid detection pipeline. Low-cost signals flag suspicious outputs, a slower verifier decides on their factual status, and unsupported answers are passed for mitigation. The red boxes denote the sections of the pipeline subject to failure.}
\label{fig:det-hybrid}
\end{figure}

No single detection strategy suffices for every setting. Human evaluation is the high-quality validation standard, and reference-based methods are strongest when trusted evidence exists. Reference-free methods scale and apply to black-box systems, but they estimate reliability only indirectly.  
Retrieval- and claim-level verification methods provide fine-grained, evidence-based judgments, but they inherit retrieval weaknesses and may fail to detect reasoning-level errors. Model based judges offer practical and scalable evaluation, although their reliability depends on the quality and independence of the evaluating model~\cite{Huang2025Survey,Ji2023Survey,AnhHoang2025Survey,Shorinwa2025UQ}.
The scientific trend is towards hybrid systems that combine obtaining uncertainty estimates with several techniques, including those that check claims, and obtain proof from other sources. In these hybrid systems, the processes of flagging suspect outputs and checking their correctness are completely separated. Moreover, uncertainty serves as a tool for controlling outputs since outputs associated with high uncertainty require retrieval, verification, repetition, or human involvement. However, this setup entails certain risks. For example, it may be that the first detection will not understand that there is a delusion, as a result, there may not be any corrective actions taken. In case the verification fails, it can authenticate incorrect output.~\cite{Pan2024Correction}. Figure~\ref{fig:det-hybrid} illustrates a hybrid detection pipeline and highlights the potential failure points during detection and verification.

The table provided in this section adduces the pros, cons, and best applications of each method in the family. Mitigation methods for flagged outputs, and the benchmarks and metrics used to evaluate the detectors themselves, are beyond the scope of this review.

\begin{table}[H]
\centering
\small
\renewcommand{\arraystretch}{1.25}
\caption{The trade-offs between the detection families}
\label{tab:det-tradeoffs}
\begin{tabular}{|L{0.15\linewidth}|L{0.23\linewidth}|L{0.25\linewidth}|L{0.20\linewidth}|}
\hline
\rowcolor{gray!15}
\textbf{Family} & \textbf{Strength} & \textbf{Main limitation} & \textbf{Best suited to} \\
\hline
Human evaluation & High semantic sensitivity; validation standard & Expensive and hard to scale; harder in specialized domains & Auditing automatic detectors \\
\hline
Reference-based, NLI & Verdict tied to explicit evidence; strong with reliable references & Depends on reference quality, completeness, and temporal validity & Grounded, source-based tasks \\
\hline
Retrieval and claim-level & Fine-grained; locates the failing statement & Inherits retrieval weaknesses; weak on reasoning-level failures & Open-domain QA, RAG, long answers \\
\hline
Reference-free: uncertainty & Needs no gold reference; model-internal and scalable & Indirect: confident models can still hallucinate & Settings where verification is costly; confabulations \\
\hline
Reference-free: consistency & Works for black-box systems & Systematic errors stay consistent & API-only access \\
\hline
LLM-as-a-Judge & Scalable; captures nuance & Judge may be biased, inconsistent, or wrong & Large-scale pipelines, validated against humans \\
\hline
Multimodal & Catches visual grounding errors & Needs modality-specific verification & MLLM outputs \\
\hline
Multilingual & Handles language-dependent patterns & Weak transfer across languages & Translation, low-resource directions \\
\hline
\end{tabular}
\end{table}

\providecolor{causecol}{RGB}{214,234,248}
\providecolor{detcol}{RGB}{220,240,220}
\providecolor{linecol}{RGB}{90,90,90}
\providecolor{conclcol}{RGB}{235,245,235}
\providecolor{edgeone}{RGB}{192,57,43}
\providecolor{edgetwo}{RGB}{230,126,34}
\providecolor{edgethree}{RGB}{142,68,173}
\providecolor{edgefour}{RGB}{22,160,133}
\providecolor{edgefive}{RGB}{121,85,72}

\providecommand{\flink}[5]{\draw[#1] ($(#2.east)+(0,#3)$) to[out=0,in=180] ($(#4.west)+(0,#5)$);}

Detection cannot be considered a single technique, as it actually represents a set of different techniques which each have their peculiarities in terms of its objective, accuracy, and use of references among some others.
There are four major aspects that can be highlighted in this part of the paper. First, the detector to be used for detection should be selected on the basis of the available evidence, not on the level of sophistication of techniques in question: if trusted reference or referenceable evidence exists, it is better to use the techniques based on evidence, and reference-free approaches should be applied only when the verification process is difficult or impossible.
Second, it is possible to say that uncertainty and consistency signals should be used as a means of navigation instead of verdict attribution. They are cheap enough to screen every output, but they are blind to errors the model repeats consistently, which are exactly the systematic errors that training-time causes produce (Section~\ref{sec:causes}) \cite{Farquhar2024Detecting}. Third, a detector's decision is only dependable as its weakest part, whether that part be old evidence, a biased judge, or a detector that uses the same system as the machine that created the evidence, thus detectors must be evaluated against human judgments and made as independent of the generator as possible. Fourth, the verdict saying only “hallucinated” provides limited information and thus detectors can be substantially better if they can determine what went wrong and how bad this is, since this information is needed for both mitigation and human investigation. Thus, we believe that our main task is not to devise one stronger detector, but rather to form pipelines of complementary signals, in which they are calibrated, aware of the evidence, and explicit about what they cannot detect while testing them in real-life situations, when shifts in domains and low-resource languages are present, and when visual data is in use instead of just in-domain data.

\section{Open Challenges \& Gaps}

Although significant strides have been made in comprehending and identifying hallucinations, there are still major issues yet to be addressed. In the previous paragraphs, we show that the issues presented here are interrelated rather than independent because causes influence all aspects of the model and its pipelines, detection methods depend on different types of evidence, and every cause that remains unsolved has a weakening effect on a specific kind of detection method, while every detection problem implies that a particular aspect has not been taken into account. As a result, a method that performs well in one setting does not necessarily remain reliable in another. The challenges below follow from this observation and are grouped into five research gaps, G1 to G5, covering origin attribution, calibrated uncertainty, generalization across domains, language and modality robustness, and the reliability of verifiers and evidence. For each gap we describe the evidence behind it and what is needed to close it.
 
\paragraph{G1: Attributing the Origin of a Hallucination}
Current detectors can often identify the presence of an error, but they cannot reliably pinpoint the stage of the pipeline where the error originated. Retrieval-augmented generation (RAG) introduces a significant complexity. In RAG, a model utilizes documents that are retrieved at query time; thus, an incorrect answer can be produced due to either missing the retrieval, retrieving documents that have only a weak connection to the answer, or using the retrieved evidence in the wrong way~\cite{Amugongo2025Retrieval,Rahman2025Hallucination}. Moreover, its output alone does not tell us which of these scenarios have occurred: be it the presence of errors in the retrieval or faulty reasoning~\cite{Li2024Drowzee}.
Claim-level verification helps identify incorrect claims but may not resolve errors arising from complex reasoning~\cite{Farquhar2024Detecting}. Apart from that, detectors can hardly distinguish random generation from systematic mistakes, and their verdicts, such as ‘supported,’ ‘contradicted,’ or ‘unresolved,’ do not systematically assess severity~\cite{Asgari2025ClinicalSafety}. The development of detectors that can report not only the source of the fault (data, knowledge, retrieval errors, reasoning problems, etc.) but also the type of error and severity of the observed fault would be helpful.

\paragraph{G2: Calibrated Uncertainty and High-Confidence Errors}
Reference-free methods estimate the reliability of model outputs without requiring an external gold reference, although their signals are indirect~\cite{Shorinwa2025UQ}. A model can produce incorrect responses with high confidence, as shown in studies of clinical reasoning~\cite{Kim2025ClinicalReasoning}, which relates to the alignment issue discussed in Section~2.3, where optimization favors producing an answer over expressing uncertainty. Semantic entropy can detect certain forms of confabulation without task-specific training labels, but it does not capture all types of hallucinations, particularly systematic errors~\cite{Farquhar2024Detecting}. Token-level uncertainty measures uncertainty over individual token choices, whereas self-consistency assesses the stability of claims across repeated generations.

\paragraph{G3: Generalization Across Domains and Tasks}
Most detectors seem to be dependent on the given domain and task. For example, a detector for factual question-answering in the open domain does not necessarily transfer to clinical reasoning or code generation \cite{Farquhar2024Detecting,Huang2025Survey}. The phenomenon of hallucination in code can depend on the developed benchmark and knowledge applied, while the context of the library may add new failure cases. In clinical reasoning, there may be cases when some benchmark has a great performance but fails in certain instances requiring cognitive flexibility~\cite{Kim2025ClinicalReasoning}. Validating by humans becomes more complex with the specialization of the domain and length of the input causing the lost-in-the-middle effect—when the information existing at the beginning and at the end is utilized more often than the one from the middle of the text~\cite{Zhang2025Leveraging}.

\paragraph{G4: Multilingual and Multimodal Robustness}
Detection research is concentrated on English text. Hallucinations are
especially prevalent in low-resource translation directions (language pairs with
little training text) and when translating out of English, and detectors
developed for English may not transfer \cite{Guerreiro2023Hallucinations}; the
data composition effects of Section~2.1 are a likely contributor. In multimodal
models, which process images alongside text, errors in object existence,
attributes, and relationships need signals that weigh visual evidence together
with the generated text, and text-only judges cannot fully assess errors that
begin in visual misperception
\cite{Yin2024MLLMsurvey,Chen2025MMHallucination}. Systems that are multilingual
and multimodal at once may face compounded grounding failures that
language-specific or modality-specific detectors barely cover.
Future detectors should therefore be evaluated across different language-resource levels and modalities, including systems that combine multilingual and multimodal inputs. 
 
\paragraph{G5: Reliable Verifiers and Temporally Valid Evidence}
The effectiveness of detection methods depends on the evidence and validation used in the process, regardless of which technique is being applied.
The problem with retrieving evidence, for example, is that the evidence may be missing, outdated, noisy, or irrelevant, leading to errors in verdicts even if the evidence exists~\cite{Amugongo2025Retrieval,Rahman2025Hallucination}.
Thus, it is important to understand how both parametric and external knowledge change over time and lead to errors in verdicts.
Another limitation arises in LLM-as-a-Judge methods, where the evaluating model may introduce its own biases and inconsistencies.
In hybrid techniques that rely on the output of cheap screening tests and a slower verification process for the results, sometimes cues may go unnoticed leading to a situation when no corrective actions are taken.
In addition, if the verification process is unreliable and LLM is responsible for generating and verifying the output at the same time, there might be a number of permanent errors.
Thus, there is a need for the use of reliable verifiers validated against human evaluation and across domains and models.

\section{Conclusion}
\label{sec:conclusion}

This research investigated what causes large language models to hallucinate and how hallucinations can be detected. Hallucination is a broad category of errors rather than a single failure mode. Hallucination can be evaluated based on the reference material (intrinsic vs extrinsic), the world knowledge that is available (contradiction vs fabrication), or following the guidelines and the model’s context (faithfulness), and these are parameters that can be broken down into error type, modality, and severity of the error. There is a wide variety of causes of hallucination, and they can be found all over the process: poor or insufficient data during training, static parametric knowledge, aiming for high likelihood instead of truth, ambiguous questions given to the model, flawed generation and misleading context, flawed retrieval, and poor vision grounding, while the domain and task amplify the disadvantages of each error. Since hallucinations arise from multiple interacting causes, no one detector can solve the problem since every method/technology solves certain parts of the issue, and the industry trend aims toward forming layered or hybrid detection pipelines.

\noindent In the deployment of LLM technologies where mistakes bear tangible consequences, the key to meaningful progress is not reliance on a more advanced detector, but rather the development of detectors that can trace the cause of an error, stand the test of time in the face of domain change, and be validated by trustworthy sources. Understanding the reasons for hallucinations remains key, as well as developing detectors that are accountable, stable across different areas and languages, and resilient to changes in language and context. Future research should therefore focus on identifying the origins of hallucinations, improving uncertainty calibration, strengthening generalization across domains and languages, developing multimodal detection capabilities, and establishing independent and reliable verification systems.

\bibliographystyle{plain}

\bibliography{refer}

\end{document}